\documentclass[twoside,11pt]{article}

\usepackage{jmlr2e}

\firstpageno{1}

\usepackage[titletoc]{appendix}
\usepackage[utf8]{inputenc} %
\usepackage[T1]{fontenc}    %
\usepackage{lmodern}
\usepackage{url}            %
\usepackage{booktabs}       %
\usepackage{amsfonts}       %
\usepackage{nicefrac}       %
\usepackage{microtype}      %
\usepackage{graphicx}
\usepackage{url}
\usepackage[usenames,dvipsnames]{color}
\usepackage[usenames,dvipsnames]{xcolor}
\usepackage{amsmath}
\usepackage{array}
\usepackage{verbatim}
\usepackage{bm}
\usepackage{subfigure}
\usepackage{wrapfig}
\usepackage{hyperref}  
\hypersetup{
	colorlinks   = true,
	citecolor    = BlueViolet
}
\newtheorem{exam}{Example}
\newtheorem{defi}{Definition}
\newtheorem{coro}{Corollary}
\newtheorem{ass}{Assumption}
\newtheorem{mylemma}{Lemma}

\RequirePackage{natbib}
\RequirePackage{graphicx}

\newcommand{\rvhilight}[1]{{\leavevmode\color{Black}#1}}
\newcommand{\rvTwohilight}[1]{{\leavevmode\color{Black}#1}}

\usepackage{lastpage}
\jmlrheading{23}{2022}{1-\pageref{LastPage}}{3/21}{4/22}{21-0218}{Song Liu, Takafumi Kanamori and Daniel J. Williams}
\ShortHeadings{Estimating Density Models with Truncation Boundaries using Score Matching}{Liu, Kanamori and Williams}

\begin{document}
\newtheorem{mytheorem}{Theorem}
\newtheorem{mylemmssu}{Lemma}
\newtheorem{prop}{Proposition}
\newtheorem{rem}{Remark}
\newcommand{\argmax}{\mathop{\rm argmax}\limits}
\newcommand{\argmin}{\mathop{\rm argmin}\limits}

\newcommand{\hilight}[1]{{#1}}
\newcommand{\highlight}[2][yellow]{\mathchoice%
  {\colorbox{#1}{$\displaystyle#2$}}%
  {\colorbox{#1}{$\textstyle#2$}}%
  {\colorbox{#1}{$\scriptstyle#2$}}%
  {\colorbox{#1}{$\scriptscriptstyle#2$}}}%

\newcommand{\unorm}[1]{\|#1\|}
\newcommand{\unorms}[1]{\unorm{#1}^2}
\newcommand{\calX}{{\mathcal{X}}}
\newcommand{\calY}{{\mathcal{Y}}}
\newcommand{\boldtheta}{{\boldsymbol{\theta}}}
\newcommand{\bolddelta}{{\boldsymbol{\delta}}}
\newcommand{\boldthetaP}{{\boldsymbol{\theta}}^{(p)}}
\newcommand{\boldthetaQ}{{\boldsymbol{\theta}}^{(q)}}
\newcommand{\boldthetaPtop}{{\boldsymbol{\theta}}^{(p)\top}}
\newcommand{\factorp}{{\phi}^P}
\newcommand{\factorq}{{\phi}^Q}
\newcommand{\boldalpha}{{\boldsymbol{\alpha}}}
\newcommand{\boldmu}{{\boldsymbol{\mu}}}
\newcommand{\boldHh}{{\widehat{\boldH}}}
\newcommand{\boldeta}{{\boldsymbol{\eta}}}
\newcommand{\bolda}{{\boldsymbol{a}}}
\newcommand{\boldH}{{\boldsymbol{H}}}
\newcommand{\boldA}{{\boldsymbol{A}}}
\newcommand{\boldp}{{\boldsymbol{p}}}
\newcommand{\boldS}{{\boldsymbol{S}}}
\newcommand{\boldK}{{\boldsymbol{K}}}
\newcommand{\boldJ}{{\boldsymbol{J}}}
\newcommand{\boldT}{{\boldsymbol{T}}}
\newcommand{\boldTheta}{{\boldsymbol{\Theta}}}
\newcommand{\boldf}{{\boldsymbol{f}}}
\newcommand{\boldu}{{\boldsymbol{u}}}
\newcommand{\boldm}{{\boldsymbol{m}}}
\newcommand{\boldone}{{\boldsymbol{1}}}
\newcommand{\boldxi}{{\boldsymbol{\xi}}}
\newcommand{\boldSigma}{{\boldsymbol{\Sigma}}}
\newcommand{\boldn}{{\boldsymbol{n}}}
\newcommand{\boldv}{{\boldsymbol{v}}}
\newcommand{\boldk}{{\boldsymbol{k}}}
\newcommand{\boldB}{{\boldsymbol{B}}}
\newcommand{\boldb}{{\boldsymbol{b}}}
\newcommand{\boldbeta}{{\boldsymbol{\beta}}}
\newcommand{\boldDelta}{{\boldsymbol{\Delta}}}
\newcommand{\nnu}{\nsample}
\newcommand{\nsample}{n}
\newcommand{\subsetr}{\boldsymbol{r}}
\newcommand{\boldthetah}{{\widehat{\boldtheta}}}
\newcommand{\mathbbR}{\mathbb{R}}
\newcommand{\mathcalD}{\mathcal{D}}
\newcommand{\trace}{\mathrm{tr}}
\newcommand{\KL}{\mathrm{KL}}
\newcommand{\numparams}{n}
\newcommand{\boldhh}{{\widehat{\boldh}}}
\newcommand{\boldh}{{\boldsymbol{h}}}
\newcommand{\Hh}{{\widehat{H}}}
\newcommand{\boldxnu}{\boldY}
\newcommand{\boldx}{{\boldsymbol{x}}}
\newcommand{\bolde}{\boldsymbol{e}}
\newcommand{\boldxp}{{\boldsymbol{x}}_{p}}
\newcommand{\boldxq}{{\boldsymbol{x}}_{q}}
\newcommand{\boldz}{{\boldsymbol{z}}}
\newcommand{\boldg}{{\boldsymbol{g}}}
\newcommand{\boldw}{{\boldsymbol{w}}}
\newcommand{\boldr}{{\boldsymbol{r}}}
\newcommand{\boldQ}{{\boldsymbol{Q}}}
\newcommand{\boldF}{{\boldsymbol{F}}}
\newcommand{\boldphi}{{\boldsymbol{\phi}}}
\newcommand{\boldzero}{{\boldsymbol{0}}}
\newcommand{\thetahat}{{\hat{\boldsymbol{\theta}}}}
\newcommand{\thetaShat}{{\hat{\boldsymbol{\theta}}_S}}
\newcommand{\thetaSchat}{{\hat{\boldsymbol{\theta}}_{S^c}}}
\newcommand{\zhat}{{\hat{\boldsymbol{z}}}}
\newcommand{\zSchat}{{\hat{\boldsymbol{z}}_{S^c}}}
\newcommand{\zShat}{{\hat{\boldsymbol{z}}_{S}}}
\newcommand{\nde}{\nsample'}
\newcommand{\boldxde}{\boldY'}
\newcommand{\boldX}{{\boldsymbol{X}}}
\newcommand{\boldY}{{\boldsymbol{Y}}}
\newcommand{\boldy}{{\boldsymbol{y}}}
\newcommand{\boldt}{{\boldsymbol{t}}}
\newcommand{\logptheta}{{\log p_\boldtheta}}
\newcommand{\logq}{{\log q}}
\newcommand{\boldYnu}{{\boldsymbol{Y}}}
\newcommand{\boldYde}{{\boldsymbol{Y}}}
\newcommand{\boldpsi}{{\boldsymbol{\psi}}}
\newcommand{\hh}{{\widehat{h}}}
\newcommand{\boldI}{{\boldsymbol{I}}}
\newcommand{\PE}{{\widehat{PE}}}
\newcommand{\ratioh}{\widehat{\ratiosymbol}}
\newcommand{\ratiosymbol}{r}
\newcommand{\ratiomodel}{g}
\newcommand{\thetah}{{\widehat{\theta}}}
\newcommand{\mathbbE}{\mathbb{E}}
\newcommand{\pnu}{p_\mathrm{te}}
\newcommand{\pde}{p_\mathrm{rf}}
\newcommand{\refsection}{\boldS_\mathrm{rf}}
\newcommand{\tesection}{\boldS_\mathrm{te}}
\newcommand{\refY}{\boldY_\mathrm{rf}}
\newcommand{\teY}{\boldY_\mathrm{te}}
\newcommand{\nseg}{n}
\newcommand{\distP}{P}
\newcommand{\distQ}{Q}
\newcommand{\iid}{\stackrel{\mathrm{i.i.d.}}{\sim}}
\newcommand{\dx}{\mathrm{d}\boldx}
\newcommand{\dy}{\mathrm{d}\boldy}

\newcommand{\gxeta}{g(\boldx;\boldeta)}
\newcommand{\Zeta}{Z(\boldeta)}
\newcommand{\Zetahat}{\hat{Z}(\boldeta)}
\newcommand{\normDtwo}{\|\nabla_{\boldtheta}^2 \ell(\boldtheta)\|}
\newcommand{\DtwoLSE}{\nabla_{\boldg}^2 \mathrm{LSE}\left(g_1, g_2, \cdots, g_{n_q}\right)}

\def\ratio{r}
\def\relratio{{\ratio}_{\alpha}}

\def\ci{\perp\!\!\!\perp} 
\newcommand\independent{\protect\mathpalette{\protect\independenT}{\perp}} 
\def\independenT#1#2{\mathrel{\rlap{$#1#2$}\mkern2mu{#1#2}}} 
\newcommand*\xor{\mathbin{\oplus}}

\newcommand{\vertiii}[1]{{\left\vert\kern-0.25ex\left\vert\kern-0.25ex\left\vert #1 
    \right\vert\kern-0.25ex\right\vert\kern-0.25ex\right\vert}}

\title{Estimating Density Models with Truncation Boundaries using Score Matching}

\author{%
  Song Liu\\ \url{song.liu@bristol.ac.uk}\\ University of Bristol\\ \AND
  Takafumi Kanamori\\
 \url{kanamori@c.titech.ac.jp} \\
  Tokyo Institute of Technology,\\  
  RIKEN AIP\\  \AND
  Daniel J. Williams\\ 
  \url{daniel.williams@bristol.ac.uk}\\
  University of Bristol
}
\date{}

\editor{}
\graphicspath{{pix/}}

\editor{Qiang Liu}

\maketitle

\begin{abstract}%
Truncated densities are probability density functions defined on truncated domains. 
They share the same parametric form  with their non-truncated counterparts up to a normalizing constant. 
Since the computation of their normalizing constants is usually infeasible, Maximum Likelihood Estimation cannot be easily applied to estimate truncated density models.
Score Matching (SM) is a powerful tool for fitting parameters using only unnormalized models. 
However, it cannot be directly applied here as boundary conditions used to derive a tractable SM objective are not
satisfied by truncated densities.
In this paper, we study parameter estimation for truncated probability densities using SM. 
The estimator minimizes a weighted Fisher divergence. 
The weight function is simply the shortest distance from a data point to the boundary of the domain. We show this choice of weight function naturally arises from minimizing the Stein discrepancy as well as upperbounding the finite-sample estimation error. 
The usefulness of our method is demonstrated by numerical experiments and a study on the Chicago crime data set. We also show that the proposed density estimation can correct the outlier-trimming bias caused by aggressive outlier detection methods.
\end{abstract}

\section{Introduction}
\begin{figure}[t]
	\centering
	\includegraphics[width=.3\textwidth]{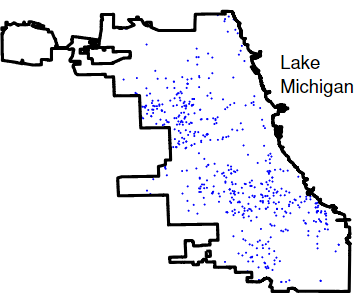}
	\caption{Boundary of Chicago, where blue dots are locations of homicides in 2008.}
	\label{fig.chicago}
\end{figure}

In many applications, we cannot observe the ``full picture'' of a problem. Instead, our window of observation is limited, so we can only observe a truncated data set. 
For example, a police department can only monitor crimes within their city's boundary, despite the fact that crimes do not automatically stop at an artificial border. Similarly, geolocation tracking data can only be observed up to the coverage of mobile signal. data sets such as these are skewed representations of actual activities due to truncation. In many cases, these truncation boundaries can be very complex. For example, the boundary of the city of Chicago is a complex polygon (see Figure \ref{fig.chicago}) 
which cannot be easily approximated by a bounding box or circle. 

The key challenge of estimating parameters in truncated densities is that the normalizing constant is not computationally tractable. The normalizing constant ensures that the integration of a density function equals to one over its input domain. As the normalization takes place in an irregular bounded domain in $\mathbbR^d$, the integration does not have a closed form in general. 
This creates a computational issue since the classic Maximum Likelihood Estimation (MLE)
requires the evaluation of such a normalizing constant. Although the normalizing constant in the likelihood function can be approximated using Monte Carlo methods \citep{Geyer1994}, it is hard to guarantee the approximation accuracy when the data is in high dimensional space and the truncation domain is complex.  

Recent years have seen a new class of estimators, called Score Matching
(SM)~\citep{Hyvaerinen2005,hyvarinen2007,Lyu2009} rise in popularity. They estimate parameters by minimizing
the Fisher-Hyv\"{a}rinen divergence~\citep{Lyu2009}. 
This divergence is defined using the difference between the gradients of log model density and log data density. The gradients are taken with
respect to the \emph{input variable}, so the normalizing constant is eliminated and is not involved in the estimation procedure. 
Thus SM is a natural candidate for estimating truncated density models. 

However, the original SM cannot work for estimating these truncated models, as the regularity condition used to derive the tractable objective function is not satisfied. 
\citet{hyvarinen2007} and \citet{YuGen2019} proposed \emph{generalized SM} to handle the distributions on the non-negative orthant $\mathbbR^d_+$. 
In generalized SM, a weight function is introduced so that the boundary condition required for deriving the tractable objective function is satisfied. 
Promising results have been observed on high-dimensional non-negative graphical model structure estimation \citep{YuGen2019}. 
When the density is truncated in a dimension-wise manner with a lower and upper bound, this problem is known as a doubly truncated distribution estimation \citep{Turnbull1976,moreira2012}. 
Though some works have tackled the problem of estimating truncated multivariate Gaussian densities \citep{daskalakis2018efficient,daskalakis19a}, little work has been done for estimating a wide range of density models with complicated truncation domains. 
\rvhilight{Note that if the truncation domain is simple, it is possible to apply the change of variable rule to our data set so that the truncated density estimation becomes a non-truncated (or doubly truncated) density estimation problem in the new domain. This technique has been widely used in modelling distributions on a sphere \citep{mardia2009directional}. However, this technique is not applicable when the truncation domain is complex such as the one in the Chicago Crime data set. }

In this paper, we proposed a novel estimator for truncated density models using generalized SM. Our choice of the weight function is the shortest distance from the data point to the truncation boundary. We show that this choice naturally arises from minimizing a Stein Discrepancy and lowering the finite sample estimation error upperbound. We also show that this distance weight function can be easily computed for many complicated domains. 

The usefulness of our method is demonstrated by numerical experiments and an application on the Chicago crime data set. Finally we apply this estimator to correct the outlier trimming bias caused by One-class Support Vector Machines. 

\section{Problem Formulation}
Denote a probability density function parameterized by $\boldtheta$ over the domain $V\subset\mathbbR^d$ as 
$p_\boldtheta(\boldx), \boldx\in V$. Without  loss of generality, we can write 
\[p_\boldtheta(\boldx) := \frac{\bar{p}_\boldtheta (\boldx)}{Z_V(\boldtheta)}, Z_V(\boldtheta) := \int_V \bar{p}_\boldtheta (\boldx) \dx,\]  
where $\bar{p}_\boldtheta$ is an unnormalized model and $Z_V(\boldtheta)$ is the normalizing constant 
so that $p_\boldtheta(\boldx)$ is integrated to $1$ over its domain $V$. The domain $V \subset \mathbb{R}^d$ may be a complicated bounded domain e.g., a polytope. 
In such cases, $Z_V(\boldtheta)$ may not have a closed form and cannot be easily evaluated.
For example, let $\bar{p}_\boldtheta$ be a Gaussian mixture model restricted in a generic polygon, then $Z_V(\boldtheta)$ does not have a closed form expression.

\textbf{Our task} is classic statistical model estimation: Suppose that $V$ is given, we want to estimate the parameter $\boldtheta$
in $p_\boldtheta(\boldx)$ using $X_q = \{\boldx_i\}_{i=1}^n,$ which is a set of observed i.i.d. samples from a truncated data generating distribution 
$Q$ with an unknown probability density function $q(\boldx), \boldx\in V$. 

\textbf{The challenge} comes from the fact that we need to obtain the estimator of $\boldtheta$ using only the unnormalized density model 
$\bar{p}_\boldtheta(\boldx)$: It is not straightforward to calculate $Z_V(\boldtheta)$ for a complicated $V$. 
Therefore, MLE, which requires evaluating the normalizing constant, cannot be performed easily. 
A popular tool for estimating unnormalized densities is Score Matching (SM) 
proposed by \citet{Hyvaerinen2005}. 
We now introduce SM and its extension~\citep{YuGen2019}, 
then explain why it cannot be readily used for estimating complicated truncated densities.

\noindent\textbf{Notation:}
The finite set $\{1,\ldots,d\}$ for a positive integer $d$ is denoted by $[d]$. 
Given a vector $\boldx$, let $x_k$ denote the $k$-th element of $\boldx$. 
Let $\langle\cdot,\cdot\rangle$ be the standard inner product, and the Euclidean norm of the vector $\bolda$ is denoted as $\|\bolda\|=\sqrt{\langle\bolda,\bolda\rangle}$. 
The $\ell^p$-norm of $\boldx$ is denoted by $\|\boldx\|_p$. Thus, $\|\boldx\|=\|\boldx\|_2$ holds.  
Let $\partial_k$ for $k\in[d]$ be the partial differential operator $\frac{\partial}{\partial x_k}$ and $\nabla_{\boldx}$ be $(\partial_{1},\ldots,\partial_{d})$
to the function $f(\boldx)$ for $\boldx\in\mathbbR^d$. 
Similarly, the gradient operator with respect to  the parameter $\boldtheta\in\Theta\subset\mathbbR^r$ of the statistical model $p_{\boldtheta}(\boldx)$ is denoted by 
$\nabla_{\boldtheta}=(\frac{\partial}{\partial\theta_1},\ldots,\frac{\partial}{\partial\theta_r})$. 
$\mathbbE_{q}[f(\boldx)]$ stands for the expectation of $f(\boldx)$ with respect to  the probability density $q(\boldx)$. 
To reduce the clutter, we sometimes shorten $p_\boldtheta(\boldx)$ and $q(\boldx)$ as $p_\boldtheta$ and $q$ wherever such abbreviations do not lead to confusion. In addition, the log-likelihood  $\log{p_{\boldtheta}}(\boldx)$ is expressed by $\ell_{\boldtheta}(\boldx)$.  $\|f\|_{L^2(Q)} := \left( \mathbbE_q\left[f(\boldx)^2\right]\right)^{1/2}$ where $Q$ denotes a distribution with density function $q(\boldx)$. Let $\bolde_i, i=1,\ldots,d$ be the unit vector along the $i$-th axis in $\mathbb{R}^d$, i.e., $\bolde_1=(1,0,\ldots,0)$, etc. We use $\overline{V}$ to represent the closure of a bounded open set $V$. 

\section{Score Matching and Its Generalization}
In this section, we introduce classic SM \citep{Hyvaerinen2005} and one of its variants \citep{YuGen2019}. 
\begin{defi} The {Fisher-Hyv\"{a}rinen~(FH) divergence \citep{Lyu2009}} between $q$ and $p_{\boldtheta}$ is defined as 
\begin{align*}
\mathrm{FH}(q,p_{\boldtheta}):= \mathbbE_{q} [\|\nabla_\boldx \log p_\boldtheta(\boldx) - \nabla_\boldx \log q(\boldx) \|^2]. 
\end{align*}
\end{defi}
Suppose $q(\boldx)>0$ on $\mathbbR^d$. Then, the FH divergence is non-negative and it vanishes if and only if $p_{\boldtheta}(\boldx)=q(\boldx)$ almost surely. 
When a density model $p_\boldtheta(\boldx)$ is defined on $V=\mathbbR^d$, SM finds an estimate of $\boldtheta$ by minimizing $\mathrm{FH}(q,p_{\boldtheta})$ over the parameter space $\boldtheta\in\Theta\subset\mathbbR^r$, i.e., 
\begin{align}
 \boldtheta_{\mathrm{SM}}
 &:= \argmin_\boldtheta \mathrm{FH}(q,p_\boldtheta) \nonumber\\ 
  \label{eq.sm0}
 &\phantom{:}
= \argmin_\boldtheta\mathbbE_{q} \left[\| \nabla_\boldx \logptheta \|^2\right] - 2\mathbbE_{q}\left[\langle \nabla_\boldx \logptheta, \nabla_\boldx \logq\rangle\right] + \mathrm{C},
\end{align}
where $C$ is a constant independent of $\boldtheta$. 
The key advantage of SM is that the normalizing constant $Z_V(\boldtheta)$ is not required when evaluating \eqref{eq.sm0} as 
$\nabla_\boldx \log{}p_\boldtheta(\boldx) = \nabla_\boldx \log \bar{p}_\boldtheta(\boldx)$. Thus, SM is widely used for estimating ``unnormalizable'' statistical models. 

Unfortunately, \eqref{eq.sm0} is not tractable as we cannot directly evaluate the second term of~\eqref{eq.sm0} without  access to $\nabla_\boldx \logq$. 
Using the integration by parts rule, however, we find that the equality, 
$$\mathbbE_{q} [\langle \nabla_\boldx \logptheta, \nabla_\boldx \logq\rangle]=
-\sum_{k=1}^{d}\mathbbE_{q}[\partial_{k}^2\logptheta]$$  
holds under the smoothness condition of $\log{p_{\boldtheta}(\boldx)}$ and $\log{q(\boldx)}$ w.r.t. $\boldx$ 
and the boundary condition 
\begin{align}
\label{eq.boundarycond}
\lim_{|x_k| \to \infty} q(\boldx) \partial_{k} \log p_\boldtheta(\boldx)=0, \forall k\in[d].
\end{align} 
Many density functions defined on $\mathbbR^d$, such as multivariate Gaussian or Gaussian mixture, satisfy these conditions. 
See \citep{Hyvaerinen2005,hyvarinen2007} for details. Thus, FH-divergence in \eqref{eq.sm0} can be re-written as 
\begin{align}
 \label{eq.sm}
 \mathrm{FH}(q,p_\boldtheta)
  =\mathbbE_{q} [\| \nabla_\boldx \log p_\boldtheta \|^2] + 2\sum_{k=1}^d\mathbbE_{q}[\partial_{k}^2\logptheta]+C
\end{align}
where the objective only relies on $q$ through the expectations which can be approximated by the empirical mean over the observed samples $X_q$. 

When $p_\boldtheta(\boldx)$ is defined on the \emph{truncated} subset in $\mathbbR^d$ such as the \emph{non-negative orthant},  \rvTwohilight{
$$\mathbbR_+^d := \{\boldx \in \mathbb{R}^d | x_k \ge 0, \forall k \in [d]\},$$ 
}the boundary condition required to apply the integration by parts rule (i.e., \eqref{eq.boundarycond}) no longer holds for many density functions such as Gaussian or Gaussian mixtures. 
To estimate parameters of density functions on the non-negative orthant, 
\citet{hyvarinen2007, YuGen2019} introduced generalized SM: 
\begin{align}
    \boldtheta_\mathrm{GSM} &:= \argmin_\boldtheta \mathrm{FH}_\boldg(q,p_\boldtheta),\nonumber\\
      \mathrm{FH}_\boldg(q,p_\boldtheta)&:= 
      \mathbbE_{q}[\|\boldg^{1/2} \circ \nabla_\boldx \log p_\boldtheta  -  \boldg^{1/2} \circ \nabla_\boldx \log q \|^2], \notag\\
      \label{eq.gen.sm}
      &= \sum_{k=1}^{d}
      \mathbbE_{q}[g_k \cdot (\partial_{k}\log p_{\boldtheta})^2] - 2\sum_{k=1}^{d} \mathbbE_{q}[g_k\cdot (\partial_{k}\log p_{\boldtheta}) ({\partial}_{k}\log q)] + C,  
\end{align}
where $\boldg(\boldx):=(g_1(\boldx),\ldots,g_d(\boldx))\in \mathbb{R}^d$ is a non-negative valued, continuously differentiable 
function with $\boldg(\boldzero)=\boldzero$. 
$\boldg^{1/2}$ is the element-wise square root operation applied on $\boldg$ and 
$\circ$ is the element-wise product. %
Examples of $\boldg$ include $g_k(\boldx)=x_k $ or $ g_k(\boldx)=\max(x_k,1), \forall k \in [d]$ given $\boldx\in\mathbbR_+^{d}$. 

The following result can be used to derive the tractable objective function from \eqref{eq.gen.sm}, which appeared in the proof of Theorem 3 in \citet{YuGen2019}. Here we restate it as a Lemma using our symbols for conveniences. 
\begin{mylemma}
  \label{lem.integralbyparts2}
 Suppose that $\log q(\boldx)$ and $\boldg(\boldx)$ are continuously differentiable almost everywhere (a.e.) on $\mathbbR^d_+$ and  $\log p_\boldtheta(\boldx)$ is 
twice continuously differentiable with respect to  $\boldx$ on $\mathbbR^d_+$. 
 Furthermore, we assume the boundary condition, \[\lim_{|x_k|\to 0+, |x_k|\to \infty} g_k(\boldx) q(\boldx) \partial_{k}\log p_\boldtheta(\boldx)=0\] for $k\in[d]$. 
 Then, 
 \[\sum_{i=1}^d\mathbbE_{q}[g_k\cdot (\partial_{k}\log p_{\boldtheta}) ({\partial}_{k}\log q)]=-\sum_{k=1}^{d}\mathbbE_{q}[\partial_{k} (g_k\partial_{k} \log p_\boldtheta)].\] 
\end{mylemma}
The proof can be found in Section A.1 in \citep{YuGen2019} which uses dimension-wise integration by parts. 
Using Lemma \ref{lem.integralbyparts2}, the generalized SM objective \eqref{eq.gen.sm} has a tractable expression, 
\begin{align}
\label{eq.gsm.obj}
  \mathrm{FH}_\boldg(q,p_\boldtheta)
    =
 \sum^d_{k=1} \mathbbE_{q}[ g_k \cdot (\partial_{k}\log p_{\boldtheta})^2] 
+ 2\sum^d_{k=1} \mathbbE_{q}[\partial_{k} (g_k\partial_{k} \log p_\boldtheta)] + C
\end{align}
where $C$ is a constant  independent of $\boldtheta$. 
One can replace the expectation with the empirical mean over $X_q$ to obtain an unbiased estimator of the above objective function. 
It is straightforward to modify the generalized SM formulation so that it works for doubly truncated distributions. 
For example, $g_k(\boldx)=\min\{x_k-a_k,b_k-x_k\}$ can be used to estimate truncated densities on the product space $\prod_{k=1}^{d}(a_k, b_k)$. 

It is worth pointing out that Lemma~\ref{lem.integralbyparts2} is a specification of the divergence theorem 
such as Green's theorem or Stokes' theorem which usually deals with a bounded domain. 
Recently, \citet{MardiaScoreMatching2016} studied an SM objective based on Stokes' theorem, for estimating densities on a Riemannian manifold. 

We intend to extend the generalized SM to a generic truncated domain. We now show the validity of the objective function \eqref{eq.gen.sm} when considering a generic domain $V$. 
\rvhilight{For two differantiable probability densities  $p(\boldx)$ and $q(\boldx)$ 
with respect to the Lebesgue measure $\mu(\cdot)$ on $V\subset \mathbb{R}^d$,  the following lemma holds:
\begin{mylemma}\label{lemma:FH0}
Suppose that $V$ is a 
\textbf{connected open subset} in $\mathbb{R}^d$ and that $g_k(\boldx)>0$ and $q(\boldx)>0, \forall \boldx \in V$. 
Then, $\mathrm{FH}_\boldg(q,p)=0$ if and only if $p=q$ for the probability density functions $p$ and $q$. 
\end{mylemma}
\begin{proof}
First, it is easy to see that $p = q \implies \mathrm{FH}_\boldg(p,q) = 0$. 
Second, 
as $\mathrm{FH}_\boldg(q,p)=0$, $g_k(\boldx)q(\boldx)\|\nabla_\boldx\log q(\boldx)-\nabla_\boldx\log p(\boldx)\|^2=0$ should hold a.e. with respect to $\mu$ on $V$.  
 Thus, we have $\nabla_\boldx(\log q(\boldx)-\log p(\boldx))=\boldzero$ 
 on $V$. 
 As $V$ is connected, 
 $\log q(\boldx)-\log p(\boldx)$ should be a constant independent of $\boldx$ on $V$. Hence, $p(\boldx)$ is proportional to $q(\boldx)$ on $V$. 
 As both are probability density functions that should be normalized, we have $p=q$. 
\end{proof}
The following theorem states that the minimizer of $\mathrm{FH}_\boldg(q,p_\boldtheta)$ is unique and is indeed the true parameter under a mild identifiability condition.
\begin{mytheorem}
\label{thm.uniqueness}
Suppose all assumptions stated in Lemma \ref{lemma:FH0} holds. 
Let $\mathcal{P}=\{p_\boldtheta(\boldx)\,|\,\boldtheta\in\Theta\subset\mathbb{R}^r\}$ be a set of parametric statistical models on 
 the connected open subset $V$ in  $\mathbb{R}^d$.
If the model $p_\boldtheta$ is identifiable in the sense that %
 $\mu(\{\boldx\,|\,p_{\boldtheta}(\boldx)\neq p_{\boldtheta'}(\boldx)\})>0$ holds for $\boldtheta\neq \boldtheta'$, and $q = p_{\boldtheta_0} \in \mathcal{P}$, then  $\argmin_{\boldtheta\in \Theta} \mathrm{FH}_{\boldg}(q,p_\boldtheta)$ is unique and is $\boldtheta_0$.  
\end{mytheorem}
\begin{proof}
 Given the assumption on $q$, we have $\mathrm{FH}_\boldg(q,p_{\boldtheta_0}) = 0$. Thus, $\boldtheta_0$ is a minimizer of $\mathrm{FH}_\boldg(q,p_{\boldtheta})$. 
  Moreover, Lemma~\ref{lemma:FH0} guarantees that $\mathrm{FH}_\boldg(p_{\boldtheta},p_{\boldtheta'})>0$ holds for $\boldtheta\neq \boldtheta'$. 
 We can prove this by the contradiction. Suppose 
 $\mathrm{FH}_\boldg(p_{\boldtheta},p_{\boldtheta'})=0$ for $\boldtheta\neq \boldtheta'$. Thus, Lemma \ref{lemma:FH0} states  
 $p_\boldtheta(\boldx)=p_{\boldtheta'}(\boldx)$ must hold on $V$. 
 This contradicts $\mu(\{\boldx\,|\,p_{\boldtheta}(\boldx)\neq p_{\boldtheta'}(\boldx)\})>0$. 
  Hence, for all $\boldtheta \neq \boldtheta_0$, $\mathrm{FH}_\boldg(p_{\boldtheta_0},p_{\boldtheta}) = \mathrm{FH}_\boldg(q,p_{\boldtheta})>0$. This leads to the conclusion that the minimizer $\boldtheta_0$ is unique. 
\end{proof}
\begin{remark}
\label{rem:non-connected}    
When the domain is not connected, $\mathrm{FH}_\boldg(q,p) = 0$ implies
that $p$ is proportional to $q$ on each connected region. 
However, for common statistical models, the proportional relationship on each connected region implies that the probability densities are the same. 
For instance, let us consider the truncated Gaussian model $p_{\boldtheta}$
on $V$ that is the disjoint union of $V_1$ and $V_2$, where
$\boldtheta$ is the mean parameter of the Gaussian distribution
$\bar{p}_{\boldtheta}$ on $\mathbb{R}^d$. 
If $p_{\boldtheta}/p_{\boldtheta_0}$ is a constant function on a small open subset in $V$, we see that $\boldtheta=\boldtheta_0$ holds. 
Hence, Theorem~\ref{thm.uniqueness} holds for such   models even when the domain $V$ is not connected. 
\end{remark} 
}
When using generalized SM on a generic domain $V$, the bigger challenge is selecting a weight function $\boldg$. 
First, we explain an intuitive way to construct $\boldg$, then we show this intuitive choice is theoretically sound and empirically \rvTwohilight{effective}.

\begin{figure}
	\centering
	\subfigure[]{
		\includegraphics[width=.25\textwidth]{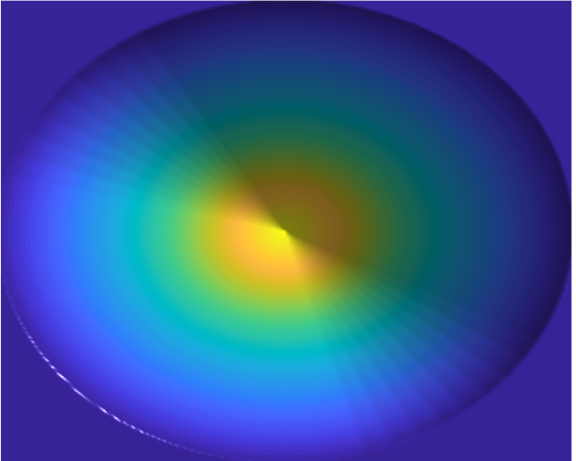}
	}
	\subfigure[]{
	\includegraphics[width=.255\textwidth]{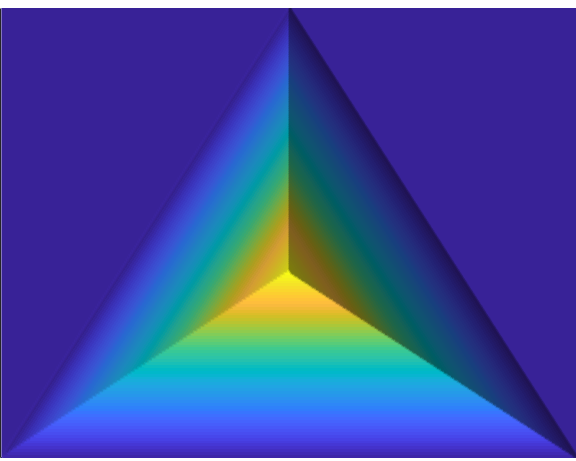}}
	\subfigure[]{
		\includegraphics[width=.25\textwidth]{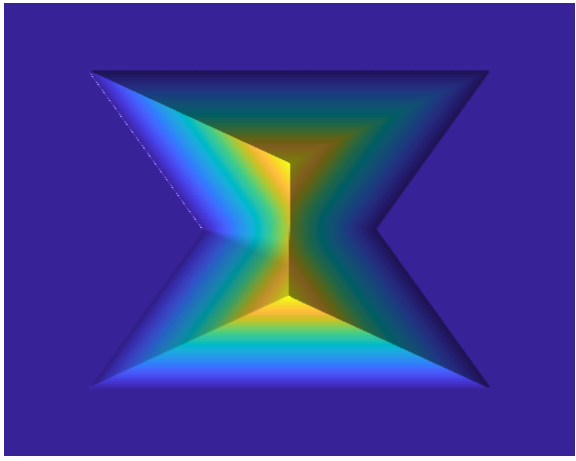}
	}
	\caption{Examples of $g_0(\boldx)$ for (a) a circular boundary (b) a triangular boundary and (c) a polygon boundary. Here $\mathrm{dist}(\cdot,\cdot)$ is the Euclidean distance function. }
	\label{fig.g0}
\end{figure}

Denote the boundary of $V$ as $\partial V$. We can design a weight function $g_k$ taking $0$ at $\partial V$, 
then hopefully an analogue to Lemma \ref{lem.integralbyparts2} would hold, giving a tractable form of the estimator. 
For example, we can consider a $\mathrm{dist}(\boldx, \boldz)$, which is a distance function between $\boldx$ and $\boldz$, and let the weight function $g_k$ be 
\begin{align}
\label{eqn:g0_distance_func}
g_k = g_0 := \min_{\boldz\in\partial{V}} 
\mathrm{dist}(\boldx, \boldz), 
\quad \forall k \in [d], ~ \boldx \in \overline{V}.
\end{align}
i.e., the distance from $\boldx$ to the truncation boundary $\partial V$.
See Figure \ref{fig.g0} for examples of $g_0$ defined on some simple bounded domains $V$. 
This weight function is intuitive and can be easily computed for many complex truncation boundaries (See Section \ref{sec:computation} for details). 

Since $g_0(\boldx)>0, \forall \boldx \in V$ by construction, Theorem \ref{thm.uniqueness} guarantees the minimizer of the generalized SM objective is the true parameter as long as $V$ is a connected open domain and $p_\boldtheta$ is correctly specified. 

However, there are a few major concerns:
\begin{enumerate}
    \item It is unclear whether 
letting $g_k = g_0$ would entail a tractable objective function.
Lemma \ref{lem.integralbyparts2} is only derived for the non-negative orthant and it assumes $g_k(\boldx)$ to be continuous differentiable a.e.. 
However, the distance weight $g_0$ is not necessarily differentiable a.e.
on $V$. 
    \item The efficacy of the distance weight $g_0$ is unclear. The weight function $g_k$ can be any function which satisfies the boundary condition ($\boldg$ taking $\boldzero$ at $\partial V$) and is positive and differentiable on $V$. It is not immediately clear why using $g_0$ as the weight function would yield good statistical estimation performance. 
\end{enumerate}

In the following sections, we address these two concerns from theoretical and empirical perspectives.
For simplicity, we refer to generalized SM with $g_k = g_0$
as truncated SM, or \emph{TruncSM} for short. 

\section{Tractable Truncated Score Matching Objective}
Theorem \ref{thm.uniqueness} states that when $V$ is a connected open domain, TruncSM is a valid density estimation criterion. 
Now we will show when $V$ is a special kind of connected open domain, TruncSM objective is computationally tractable. 

The key step of deriving a tractable SM objective is to show that the distance weight $g_0$ is \emph{weakly differentiable}, so an analogue to Lemma \ref{lem.integralbyparts2} can be proven. 
Let us formally define the notion of a weakly differentiable function via Sobolev-Hilbert space:
\begin{defi}[Sobolev-Hilbert space]: 
	Let $L^2(V)$ be the $L^2$ space on $V\subset\mathbbR^d$ endowed with the Lebesgue measure. 
	Then, $H^{1}(V)$ is the Sobolev-Hilbert space space defined by 
	\begin{align*}
		H^{1}(V) = \bigg\{ f\in L^2(V)\,\bigg|\,\|f\|_{L^2(V)}^2+\sum_k \|D_{k}f\|_{L^2(V)}^2 <\infty\bigg\}, 
	\end{align*}
	where $D_{k}$ is the weak derivative corresponding to $\partial_{k}$ and $\|f\|_{L^2(V)} = \sqrt{\int_V |f(\boldx)|^2 \dx}$.
\end{defi}

In this paper, we focus on a special type of open and connected domain, called \emph{Lipschitz domain}. 
Intuitively speaking, a Lipschitz domain $V\subset \mathbbR^d$ is a bounded connected open domain whose local boundary is a level set of some Lipschitz function. In engineering applications, most domains are Lipschitz domains. However, it does not include domains with ``cracks'' (see examples below). \rvTwohilight{All polytopes are Lipschitz domains.} \hfill\break
\begin{wrapfigure}{r}{0.2\textwidth}
  \begin{center}
  \subfigure[\rvTwohilight{Lipschitz domain}]{
    \includegraphics[width=0.18\textwidth]{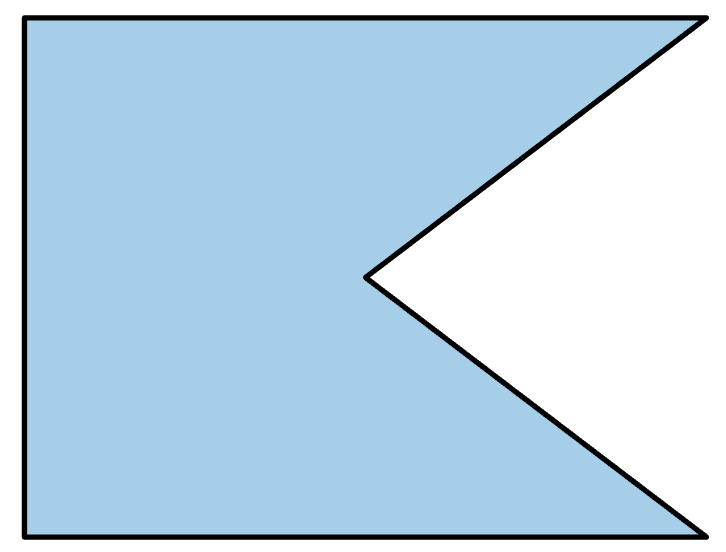}
    }
  \subfigure[\rvTwohilight{Crack domain}]{
    \includegraphics[width=0.18\textwidth]{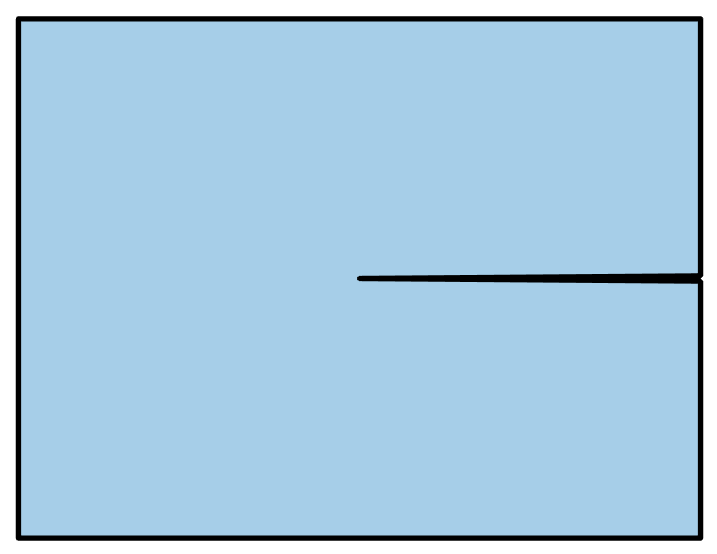}  }
  \end{center}
\vspace*{-2cm}
\end{wrapfigure}

\begin{defi}[Lipschitz Domain]
	Let $V$ be an open and bounded domain in $\mathbbR^d$. We say $V$ is a \textbf{Lipschitz domain} if 
	for any $\boldx\in\partial{V}$, there exists an $r>0$ and a Lipschitz function $f(x_1,\dots,x_{d-1})$ such that 
	$V\cap B(\boldx,r)$ is expressed by $\{\boldz \in B(\boldx,r)\,|z_d>f(z_1,\dots,z_{d-1})\,\}$
	upon a transformation of the coordinate system if necessary.  $B(\boldx,r)$ denotes a $d$-dimensional ball centered at $\boldx$ with radius $r$. 
\end{defi}
The full definition of the weak derivative and more details about Lipschitz domains and Sobolev-Hilbert spaces can be found in Section 7 of \citep{atkinson2005theoretical}. 
We prove the following lemma which states that 
 $g_0$ defined on a \emph{Lipschitz domain} is in $H^1(V)$, thus is \emph{weakly differentiable}.  The proof can be found in Appendix \ref{appendix:lemma2}. 
\begin{mylemma}
	\label{lem.2} Suppose $V$ is a Lipschitz domain, then
	$g_0 \in H^{1}(V)$. 
\end{mylemma}

The classic Green's theorem or Stokes' theorem used in Lemma~\ref{lem.integralbyparts2} cannot be applied to derive a tractable objective anymore as functions in $H^1(V)$ are \emph{not} differentiable in a classic sense. 
However, there exists an extension of Green's theorem of weakly differentiable functions (Proposition 7.6.1 \citep{atkinson2005theoretical}).

\begin{mylemma}
	[Extended Green's Theorem]
	\label{thm.greenI}
	For the Lipschitz domain $V\subset\mathbbR^d$, suppose $f_1, f_2 \in H^{1}(V)$, 
	\begin{align*}
		\int_{V} f_1(\boldx)\partial_{k} f_2(\boldx)  d\boldx
		= \int_{\partial{V}} f_1(\boldx) f_2(\boldx) \nu_k(\boldx) ds
		- \int_{V} f_2(\boldx) \partial_{k} f_1(\boldx) d\boldx, \quad \forall k\in[d],
	\end{align*}
	where $(\nu_1,\ldots,\nu_d)$ is the unit outward normal vector on $\partial{V}$ and $ds$ is the surface element on $\partial{V}$. 
\end{mylemma}

Now we apply Lemma \ref{lem.2} and \ref{thm.greenI} to obtain a tractable form of TruncSM objective:
\begin{mytheorem}
	\label{theorem:general_objfunction}
	Assume $V \subset \mathbb{R}^d$ is a Lipschitz domain.
	Suppose  $ q, \partial_{k}\log p_\boldtheta\in H^1(V)$ and
	that for any $\boldz\in\partial{V}$ it holds that 
	\[
    \lim_{\boldx\!\rightarrow\boldz} q(\boldx) g_0(\boldx)\partial_{k}\log p_\boldtheta(\boldx)\nu_k(\boldz) =0, \forall k \in [d], 
	\]
    where $\boldx\rightarrow\boldz$ takes any point sequence converging to $\boldz\in\partial{V}$ into account.
	Then, we have 
 \[\sum_{i=1}^d\mathbbE_{q}[g_0\cdot (\partial_{k}\log p_{\boldtheta}) ({\partial}_{k}\log q)]=-\sum_{k=1}^{d}\mathbbE_{q}[\partial_{k} (g_0\partial_{k} \log p_\boldtheta)].\] 
\end{mytheorem}

The proof of the above theorem can be found in Appendix~\ref{appendix:Proof_Trunc-SM-tractable}. 
This indicates that TruncSM indeed has a tractable objective function.
\begin{align}
\label{eq.tractable}
	\boldtheta_\mathrm{TSM} &:= \argmin_\boldtheta \mathrm{FH}_{g_0}(q,p_\boldtheta),\nonumber\\
	&= \argmin_\boldtheta \sum_{k=1}^{d}
	\mathbbE_{q}[g_0 \cdot (\partial_{k}\log p_{\boldtheta})^2] + 2\sum_{k=1}^{d}\mathbbE_{q}[\partial_{k} (g_0\partial_{k} \log p_{\boldtheta})]\notag \\
	&= \argmin_\boldtheta \sum_{k=1}^{d}
	\mathbbE_{q}[g_0 \cdot (\partial_{k}\log p_{\boldtheta})^2] + 2\sum_{k=1}^{d}\mathbbE_{q}[ g_0\partial^2_{k} \log p_{\boldtheta}] + 2\sum_{k=1}^{d}\mathbbE_{q}[\partial_{k} g_0 \cdot \partial_{k} \log p_{\boldtheta}],
\end{align}
where each expectation can be approximated using samples from the data set $X_q \sim Q$. 

\rvhilight{Although \eqref{eq.tractable} is similar to the generalized SM objective \eqref{eq.gsm.obj} (it replaces $g_k$ with $g_0$ ), this result cannot be taken for granted. It highlights an important restriction of TruncSM: $V$ needs to be a Lipschitz domain, and this constraint is required to ensure the weak differentiability of $g_0$, and is connected with the Lipschitzness of $g_0$. This result also suggests that we may be able to bypass this constraint when using different weight functions in generalized SM. The study along this line can be an interesting future work. }

Now we turn our focus to the efficacy of TruncSM. In the next section, we show TruncSM is a Minimum Stein Discrepancy Estimator \citep{BarpMinimumStein2019}. 

\section{Truncated Score Matching as Minimum Stein Discrepancy Estimator}
\label{subsec:TruncatedScoreMatching_MSD}

Maximum Stein Discrepancies are a family of discrepancies which measure the differences between two distributions \citep{Chwialkowski2016,LiuQ2016}. For simplicity, we assume that $q$ and $\ell_\boldtheta$ are smooth in this section. 
\begin{defi}
 Given an $\boldf: \mathbb{R}^d \to \mathbb{R}^d$, a \textbf{Stein operator} is defined as 
 $$T_{p} \boldf := \sum_{k=1}^{d} \left\{(\partial_k \log p) \cdot f_k + \partial_k f_k\right\},$$
 where $\boldf=(f_1,\ldots,f_d)$, $p$ is a probability density function and $\mathcal{S}_p := \{\boldf\,|\,\mathbb{E}_{p}[T_{p} \boldf] = 0\}$ is called a \textbf{Stein class} of $p$.
The Maximum Stein Discrepancy between two densities $q$ and $p$ \citep{Chwialkowski2016,LiuQ2016} is defined as $\max_{\boldf\in\mathcal{S}_p} \mathbbE_q \left[T_p \boldf\right]$.
\end{defi}
By constructing different Stein classes $\mathcal{S}_p$, we obtain different Maximum Stein Discrepancies.

Since our task is estimating a parametric density model $p_\boldtheta$, we can consider a 
density estimator that minimizes this discrepancy, i.e., $\argmin_\boldtheta\max_{\boldf\in\mathcal{S}_{p_\boldtheta}} \mathbbE_q \left[T_{p_\boldtheta} \boldf\right]$. This estimator has been shown to be effective, robust, and closely related to SM. It is called Minimum Stein Discrepancy Estimator \citep{BarpMinimumStein2019}.

When using the above estimator, the key issue is constructing a Stein class $S_{p_\theta}$ which is expressive enough to capture subtle differences between $q$ and $p_\boldtheta$. In the following theorem, we construct a Stein class with $\boldf$ that is the product of a smooth function and a Lipschitz function. We show the Maximum Stein Discrepancy using this specific Stein class becomes FH divergence weighted by the distance weight $g_0$.  

First, we introduce a Lemma whose proof can be found in Appendix \ref{sec.minmax.SM}.

\begin{mylemma}
\label{lem.minmax.SM}
Let $\mathrm{Lip}^L_0(V)$ be the set of all functions $f: \overline{V} \to \mathbbR$   
\begin{itemize}
    \item that are $L$-Lipschitz continuous with respect to $\mathrm{dist}(\cdot, \cdot)$
    \item satisfies the property that  $f(\boldx)=0, \forall \boldx\in \partial V$
\end{itemize}
then
$\max_{g_k \in {\mathrm{Lip}^L_0}, ~ \forall k \in [d]} \mathrm{FH}_{\boldg}(q,p_\boldtheta)
=
L\cdot \mathrm{FH}_{g_0}(q,p_\boldtheta).$
\end{mylemma}
\begin{mytheorem}
\label{thm.stein.dis}
 Let $\mathcal{F}$ be a function class such that $\forall \boldf\in \mathcal{F}$,
 $\boldf = \boldh \circ \boldg^{1/2}$ and 
 $\mathbb{E}_q \|\boldh(\boldx)\|^2  \le 1$, 
 where $\boldh: \overline{V} \to \mathbbR^d$ is a smooth function and 
 $\forall k, g_k \in \mathrm{Lip}^L_0(V)$
 then
 \begin{itemize}
 	\item $\mathcal{F}$ is a Stein class of a smooth density $q$ defined on $V$. 
 	\item If $\partial_k \ell_\boldtheta$ is smooth with respect to  $\boldtheta$ for all $k$,  
 	we have 
 	$\max_{\boldf \in \mathcal{F}} \mathbbE_q \left[T_{p_\boldtheta} \boldf\right]=\sqrt{L\cdot\mathrm{FH}_{g_0}(q,p_\boldtheta)}$.
 \end{itemize}
\end{mytheorem}
The proof can be found in Appendix~\ref{appendix:TruncSM_MSD} and it is partly based on the proof of
Theorem 2 in \citet{BarpMinimumStein2019}.
Theorem \ref{thm.stein.dis} shows that the TruncSM objective is a Maximum Stein Discrepancy thus $\boldtheta_\mathrm{TSM}$ is a Minimum Stein Discrepancy Estimator. 

Similarly, we can show how a ``capped'' distance weight arises by choosing a slightly different family for $g_k$ in Theorem \ref{thm.stein.dis}. We find this capped weight function can be also effective in many tasks (see Section \ref{sec:case_study} and \ref{sec.weightfunc}).  
\rvTwohilight{
\begin{coro}
\label{col.capped}
Let us define $\overline{\mathrm{Lip}}^L_0(V):= \{f\in \mathrm{Lip}^L_0(V)|f(\boldx)\le 1\}$ and a new function class $\overline{\mathcal{F}}$ such that for all $\boldf \in \overline{\mathcal{F}}$, $\boldf = \boldh \circ \boldg^{1/2}$,
 $\mathbb{E}_q \|\boldh(\boldx)\|^2  \le 1$, 
and 
 $g_k \in \overline{\mathrm{Lip}}^L_0(V)$, then
\begin{align*}
\max_{\boldf \in \overline{\mathcal{F}}} \mathbbE_q \left[T_{p_\boldtheta} \boldf\right]
=
\sqrt{\mathrm{FH}_{\bar{g}_L}(q,p_\boldtheta)}, 
\end{align*}
where $\bar{g}_L = \min(1,L \cdot g_0)$.
\end{coro}
}
The proof can be found in Appendix~\ref{sec.col.1}. 

\citet{YuGen2019} proposed to use a weight function 
$g_k = \min(1,x_k), k=1,\ldots,d$ 
for estimating density functions defined on the non-negative orthant. 
This weight function is the special case of $\bar{g}_L$ when $V$ is the entire non-negative orthant. 
We refer to $\bar{g}_L$ as a ``capped weight function''. 
The capped weight function is discussed in the recent paper by \citet{yu2020generalized} in the context of generalized SM for the unbounded $V$. 

\rvTwohilight{
Although Theorem \ref{thm.stein.dis} and Corollary \ref{col.capped} show that $g_0$ and $\bar{g}_L$ are natural in the sense that they both give rise to Stein divergences, these choices may not be necessarily efficient in statistical inference tasks. In Section \ref{sec:EstimationErrorBound} and \ref{sec:Experiments}, we show that our choices of  weight functions also lead to good statistical inference performance. }

In the following section, we investigate the statistical guarantee of TruncSM through finite sample estimation error analysis.

\section{Finite Sample Statistical Guarantee  of Generalized Score Matching and Optimality of Truncated Score Matching}
\label{sec:EstimationErrorBound}
To discuss the efficiency of the TruncSM estimator, we first establish a finite sample estimation error bound for generalized SM. Using this result, we show that the TruncSM estimator is a good statistical estimator. 
There have been studies on the \emph{asymptotic accuracy} of generalized SM in, for example, \citep{YuGen2019} and \citep{BarpMinimumStein2019}. 
However, it is hard to study the impact of weight functions  
from the asymptotic variance due to its complicated expression.

We now establish an estimation error bound for generalized SM using a generic weight function $\boldg$. 
For the convenience of discussion in this section, let us rewrite the generalized SM objective function as 
\begin{align}
	\label{eqn:estfunc_TruncSM}
	M(\boldtheta) = \mathbb{E}_q[m_\boldtheta(\boldx)], \text{ where }
	m_{\boldtheta}(\boldx):=\sum_{k=1}^{d} \left\{
	\big[\underbrace{{(\partial}_{k}\ell_{\boldtheta})^2   +  2\partial_{k}^2\ell_{\boldtheta}}_{=:A_k(\boldx;\boldtheta)}  \big]
	g_k
	+
	\underbrace{2\partial_{k}\ell_{\boldtheta}}_{=:B_k(\boldx;\boldtheta)}\partial_{k} g_k \right \},
\end{align}
where we abbreviated $\log p_\boldtheta$ as $\ell_\boldtheta$.
It holds that $\mathrm{FH}_{\boldg}(q,p_\boldtheta)=M(\boldtheta)+C$ in which $C$ is independent of $\boldtheta$. 
Given finite samples $X_q \sim Q$,  $M(\boldtheta)$ is approximated by the empirical mean:
\begin{align*}
	\hat{M}(\boldtheta)=\frac{1}{n}\sum_{i=1}^{n}m_{\boldtheta}(\boldx_i). 
\end{align*}
Then $\hat{\boldtheta}$, 
the minimizer of $\hat{M}(\boldtheta)$ s.t. $\boldtheta\in\Theta\subset\mathbbR^r$ is an M-estimator~\citep{vaart00:_asymp_statis} with 
the estimation function $m_{\boldtheta}(\boldx)$. 
\rvhilight{
If $\ell_\boldtheta$ is the log-likelihood of an exponential family distribution, i.e., $\ell_\boldtheta := \boldtheta^\top \boldt(\boldx) - \log Z(\boldtheta)$, where $Z(\boldtheta)$ is the normalizing constant,  
\[
\hat{M}(\boldtheta) = \frac{1}{n}\sum_{i=1}^{n}\sum_{k=1}^{d} \left\{ \left[ \boldtheta^\top  \partial_k \boldt(\boldx_i) \partial_k \boldt(\boldx_i)^\top \boldtheta + 2 \boldtheta^\top \partial^2_k\boldt(\boldx_i) \right]g_k(\boldx_i) + 2  \boldtheta^\top  \partial_k \boldt(\boldx_i) \partial_k g_k  \right\},
\]
which is convex and quadratic with respect to $\boldtheta$. Thus, a unique $\hat{\boldtheta}$ can be easily obtained as long as $\frac{1}{n}\sum_{i=1}^{n}\sum_{k=1}^{d} g_k(\boldx_i)  \partial_k \boldt(\boldx_i) \partial_k \boldt(\boldx_i)^\top \in \mathbb{R}^{r\times r}$ is invertible. However, our theorem below does not assume that $p_\boldtheta$ has a specific form. }

\subsection{Non-Asymptotic Error Bound} 
We first establish the non-asymptotic error bound of the generalized SM procedure~\eqref{eqn:estfunc_TruncSM}.
Let $\boldtheta^*\in\Theta$ be the minimizer of $M(\boldtheta)$ over $\Theta$. 
We assume that the optimal parameter $\boldtheta^*$ is well-separated from other neighbouring parameters in terms of the population objective values:
\begin{ass}
\label{ass.sep}
Assume that there exists $\alpha>1$ such that 
\begin{align}
 \label{eqn:ass1}
 \inf_{\boldtheta: \|\boldtheta-\boldtheta^*\|\ge\delta}M(\boldtheta)-M(\boldtheta^*)\geq C_{\boldg} \delta^\alpha
\end{align}
 holds for any small $\delta>0$. 
 Here, $C_{\boldg}$ is a positive constant that depends on the weight function $\boldg$
 such that $C_{a\boldg}=a C_{\boldg}$ for any positive constant $a$.
\end{ass}
\rvhilight{
Although we mainly focus on the dependency between the convergence rate and $\boldg$, the constant $C_\boldg$ can also depend on the dimensionality of the parameter space $r$, as we demonstrate in the following example.

\begin{exam}
\label{exam:exp-family}
Let us consider the exponential family
    \begin{align}
    \label{eq.exp.fami}
        p_{\boldtheta}(\boldx)=\exp\left\{\sum_{k=1}^{r}t_k(\boldx)\theta_k-\phi(\boldtheta)\right\},\, \boldtheta\in\Theta\subset\mathbb{R}^r
    \end{align}
such that $\Theta$ is bounded, i.e., $\|\boldtheta\|< R$. 

Suppose $q=p_{\boldtheta^*}$, guaranteed by Theorem \ref{thm.uniqueness} under mild conditions. 
Then, some calculation yields that $M(\boldtheta)$ is a convex quadratic function, 
\begin{align*}
 M(\boldtheta)-M(\boldtheta^*) = \left(\boldtheta-\boldtheta^*\right)^\top\left(\sum_{k=1}^d T_{k}\right)(\boldtheta-\boldtheta^*), 
\end{align*}
where $(T_k)_{ij}=(\mathbb{E}_{\boldtheta^*}[g_k(\boldx)\partial_k t_i(\boldx) \partial_k t_j(\boldx)])_{ij}\in\mathbb{R}^{r\times r}$. Note that 
$T_k$ is positive semidefinite. 
We assume that $\frac{1}{d}\sum_{k=1}^d T_{k}\succeq \bar{T}\succ O$ holds for a positive definite matrix 
$\bar{T}\in\mathbb{R}^{r\times r}$, where the inequalities are defined in the sense of positive definiteness. 
A mild assumption is that the eigenvalues of $\bar{T}$ does not depend on $d$. 
Let $\lambda_1\geq \cdots\geq \lambda_r>0$ be eigenvalues of $\bar{T}$, then, we have 
\begin{align}
\label{eq.assump1}
 M(\boldtheta)-M(\boldtheta^*) \geq d \lambda_r \|\boldtheta-\boldtheta^*\|^2. 
\end{align}
Hence, $C_\boldg = d \lambda_r$ and $\alpha=2$ meet Assumption 1. One can also confirm that $C_{a\boldg} = aC_{\boldg}$ for any $a>0$. 
\end{exam}
}
\begin{ass}
\label{ass.thetaconsistency}
 The sequence $\hat{\boldtheta}$ converges to $\boldtheta^*$ 
 in probability.
\end{ass}
 The consistency of $M$-estimator has been well-studied in the statistical literature, thus we do not discuss this in detail. Here we are only interested in the rate that governs the convergence of $\hat{\boldtheta}$ and its implication on choosing $g_k$. A set of sufficient conditions for proving the consistency of $\hat{\boldtheta}$
is discussed in Theorem 5.7 of \citet{vaart00:_asymp_statis}. 

We also make continuity assumptions on $m_\boldtheta(\boldx)$. 
This is needed to ensure the upperboundedness of the covering number when proving the convergence rate.
\begin{ass}
\label{ass.count}
 For the function $A_k$ and $B_k$ in $m_\boldtheta(\boldx)$, there exists $\dot{A}_k,\dot{B}_k, \forall k$ such that  
\begin{align}
\label{eqn:dotA_and_dotB}
\begin{array}{l}
 |A_k(\boldx,\boldtheta_1)-A_k(\boldx,\boldtheta_2)| \leq 
  \dot{A}_k(\boldx)\|\boldtheta_1-\boldtheta_2\|,\\
 |B_k(\boldx,\boldtheta_1)-B_k(\boldx,\boldtheta_2)| \leq 
 \dot{B}_k(\boldx)\|\boldtheta_1-\boldtheta_2\|. 
\end{array}
\end{align}
\end{ass}
If $A_k(\boldx, \boldtheta)$ is differentiable with respect to $\boldtheta$ for all $k$, due to Taylor expansion and Schwarz inequality, we see $|A_k(\boldx, \boldtheta_1) - A_k(\boldx, \boldtheta_2)| \le \sup_{\boldtheta\in \Theta} \|\nabla_\boldtheta A_k(\boldx,\boldtheta)\| \cdot \| \boldtheta_1 - \boldtheta_2 \|$.
Applying the same argument to $B_k(\boldx, \boldtheta)$, we
can see that both $\dot{A}_k:=\sup_{\boldtheta\in\Theta}\|\nabla_{\boldtheta} A_k(\boldx,\boldtheta)\|_2$ and $\dot{B}_k:=\sup_{\boldtheta\in\Theta}\|\nabla_{\boldtheta} B_k(\boldx,\boldtheta)\|_2$ would satisfy Assumption \ref{ass.count}. 

Let us define a function $\Gamma(\boldg;A,B)$ using $\dot{A}_k$ and $\dot{B}_k$:
\begin{align}
\label{eqn:Gamma_g1}
 \Gamma(\boldg;A,B):=
 \sum_{k=1}^{d}  
    \bigg\{
    (\mathbb{E}_{q}[\dot{A}_k^4]\mathbb{E}_{q}[g_k^4])^{1/4}
 +  (\mathbb{E}_{q}[\dot{B}_k^4]\mathbb{E}_{q}[(\partial_{k}g_k)^4])^{1/4}
  \bigg\},
\end{align}
then we have the following non-asymptotic estimation error bound of generalized SM: 
\rvhilight{
\begin{mytheorem}
\label{theorem:convergence_GSM}
 Suppose that Assumptions \ref{ass.sep}, \ref{ass.thetaconsistency} and  \ref{ass.count} hold and that $g_k, \partial_{k}g_k$, $\dot{A}$ and $\dot{B}$ have the fourth order moment under 
 the population distribution $q$.
 Then for 
$ \delta < CK_\alpha \cdot\frac{ \sqrt{r}}{2^{\alpha-1}} \frac{\Gamma(\boldg; A, B)}{C_\boldg } $
 we have 
 \begin{align}
 \label{eq.convergencebound}
  P\left[
  \|\hat{\boldtheta}-\boldtheta^*\|\le
    \left(C K_{\alpha} \cdot \frac{ \Gamma(\boldg; A,B)}{\delta C_{\boldg}} \cdot\sqrt{\frac{r}{n}}\right)^{1/(\alpha-1)}
    \right]
    \ge 1-\delta, 
  \end{align}
where $C$ is a universal constant and $K_{\alpha}=\frac{2^{2\alpha}}{2^{\alpha-1}-1}$.
\end{mytheorem}
}
The proof can be found in Appendix \ref{appendix:convergence_GSM}. 
In the proof, we use the convergence analysis of the M-estimator according to Section~5.8 in \citet{vaart00:_asymp_statis}. 
\rvhilight{
\begin{exam}
  Let us analyse the estimation error for the exponential family. 
Following the model definition  \eqref{eq.exp.fami}, we can see that $\alpha=2$ and $\Gamma_\boldg = O(d)$\footnote{For simplicity, let us shorten ${\Gamma(\boldg; A,B)}$ as $\Gamma_\boldg$ from now on. }.
The analysis in Example~\ref{exam:exp-family} leads to  $C_\boldg=d \lambda_r$. Thus, %
we have $\frac{\Gamma_\boldg}{C_\boldg} \simeq \frac{1}{\lambda_r}$ and 
\begin{align*}
 \Pr\bigg(
 \|\widehat{\boldtheta}-\boldtheta^*\|_2\leq \frac{C'}{\delta \lambda_r}\sqrt{\frac{r}{n}}
 \bigg)\geq 1-\delta, 
\end{align*}
where $C'$ is a constant independent of $d,r$ and $\delta$. 
In the above bound, the probability $\delta$ and the sample size $n$ are variables and 
other parameters such as $r$ and $\lambda_r$ are regarded as a fixed constant. 

\end{exam}
}

Theorem \ref{theorem:convergence_GSM} shows how the choice of weight $\boldg$ affects the convergence rate of $\|\hat{\boldtheta}-\boldtheta^*\|_2$. 
The only term involving $\boldg$ on the RHS of \eqref{eq.convergencebound} is $\frac{\Gamma_\boldg}{C_{\boldg}}$. 
Naturally, we would like to choose a $\boldg$ such that $\frac{\Gamma_\boldg}{C_{\boldg}}$ is minimized.
Some details will be discussed in Section~\ref{subsec:Choice_WeightFunc} and Appendix \ref{sec.capped.distance}.

\subsection{Choice of Weight Function}
\label{subsec:Choice_WeightFunc}
\rvhilight{
When our model $p_\boldtheta$ is not correctly-specified, $\boldtheta^*$ depends on the weight function $\boldg$. For simplicity, we assume the model is correctly specified and is identifiable, thus Theorem \ref{thm.uniqueness} ensures that $\boldtheta^*$ is unique and $p_{\boldtheta^* }= q$ under mild conditions.
}

We cannot minimize  $\frac{\Gamma_\boldg}{C_{\boldg}}$ with respect to $\boldg$ analytically, as we do not know $q(\boldx)$ and the closed form expressions of $\dot{A}_k$ or $\dot{B}_k$. Numerical minimization of $\frac{\Gamma_\boldg}{C_{\boldg}}$ can also be cumbersome. Instead, we present a lightweight selection procedure, which eventually gives rise to the distance weight function $g_0$ by controlling an upperbound of $\frac{\Gamma_\boldg}{C_{\boldg}}$. 

Suppose there exists a constant $\Gamma_\mathcal{G}\ge \sup_{\boldg \in \mathcal{G}}\Gamma_\boldg$, where $\mathcal{G}$ is a function family from which $\boldg$ is chosen. We can obtain an upperbound $\frac{\Gamma_\mathcal{G}}{C_{\boldg}} \ge \frac{\Gamma_\boldg}{C_{\boldg}}$. 
The choice of $\mathcal{G}$ is important as $\Gamma_\mathcal{G}$ may not exist or be very large for some $\mathcal{G}$. However, \eqref{eqn:Gamma_g1} suggests that as long as $|\partial_k g_k|$ and $|g_k|$ are upperbounded and $\dot{A}_k$ and $\dot{B}_k$ are well behaved, $\Gamma_\mathcal{G}$ should exist.
Let us consider 
\[\mathcal{G}:=
\{\boldg: \overline{V}\rightarrow\mathbb{R}^d 
| g_k \in \mathrm{Lip}^1_0(V), \forall k \in [d] \}.\]  We can see $\forall \boldg \in \mathcal{G}, |\partial_k g_k| \le 1$ due to the property of Lipschitz function and $|g_k|$ is bounded as it is defined over a bounded domain. 

Note that we have used Lipschitz functions to construct a
Stein class in Section \ref{subsec:TruncatedScoreMatching_MSD}, but here, Lipschitz functions emerge from a different context: Suppressing the upperbound of $\Gamma_\boldg$.

After fixing $\mathcal{G}$, we seek to  maximize the denominator $C_\boldg$ by choosing an appropriate $\boldg\in \mathcal{G}$. 
Let us introduce the following proposition: 
\begin{prop}
\label{prop.optim}
Given the $\mathcal{G}$ defined above, suppose $q = p_{\boldtheta^*}$, then 
\begin{align*}
 \inf_{\boldtheta: \|\boldtheta-\boldtheta^*\|\ge\delta}M_{g_0}(\boldtheta)-M_{g_0}(\boldtheta^*) \geq \sup_{\boldg \in \mathcal{G}} \inf_{\boldtheta: \|\boldtheta-\boldtheta^*\|\ge    \delta}  M_{\boldg}(\boldtheta)-M_{\boldg}(\boldtheta^*), 
\end{align*}
where $M_{\boldg}(\boldtheta)$ is $M(\boldtheta)$ using $\boldg$ as weight function. 
\end{prop}
\begin{proof}
\begin{align*}
 &\inf_{\boldtheta: \|\boldtheta-\boldtheta^*\|\ge\delta}M_{g_0}(\boldtheta)-M_{g_0}(\boldtheta^*) \\
 =&\inf_{\boldtheta: \|\boldtheta-\boldtheta^*\|\ge\delta}\mathrm{FH}_{g_0}(\boldtheta) - \mathrm{FH}_{g_0}(\boldtheta^*)\\
 =& \inf_{\boldtheta: \|\boldtheta-\boldtheta^*\|\ge\delta} \sup_{\boldg \in \mathcal{G}} \mathrm{FH}_{\boldg}(\boldtheta) - \mathrm{FH}_{\boldg}(\boldtheta^*)\\
 =& \inf_{\boldtheta: \|\boldtheta-\boldtheta^*\|\ge\delta} \sup_{\boldg \in \mathcal{G}} M_{\boldg}(\boldtheta)-M_{\boldg}(\boldtheta^*)
 \geq \sup_{\boldg \in \mathcal{G}} \inf_{\boldtheta: \|\boldtheta-\boldtheta^*\|\ge    \delta}  M_{\boldg}(\boldtheta)-M_{\boldg}(\boldtheta^*).
\end{align*}
The second equality is due to Lemma \ref{lem.minmax.SM} and  $\mathrm{FH}_{\boldg}(\boldtheta^*) \equiv 0, \forall \boldg$.
The inequality is due to the max-min inequality.
\end{proof}
Proposition \ref{prop.optim} shows, 
when using the distance weight $g_0$ as the weight function, Assumption \ref{ass.sep} should hold for a constant $C_{g_0} = \sup_{\boldg\in \mathcal{G}} C_{\boldg}$.
Since $\boldg_0 = (g_0, \dots, g_0)$ is in $\mathcal{G}$ (see the proof of Lemma \ref{lem.2}), $g_0$ maximizes $C_\boldg$ for all $\boldg\in \mathcal{G}$. 

As $\Gamma_\mathcal{G}$ does not depend on any individual $\boldg$, $g_0$ minimizes the upper bound  $\frac{\Gamma_\mathcal{G}}{C_{\boldg}}$.

Note that we can reach a similar conclusion by replacing $\mathcal{G}$ with $\overline{\mathcal{G}}:=\{\boldg \in \mathbb{R}^d | g_k \in \overline{\mathrm{Lip}}^1_0(V), \forall k \in [d] \}$ and the minimizer of $\frac{\Gamma_\mathcal{G}}{C_{\boldg}}$ would change from distance $g_0$ to
the capped distance $\bar{g}_L$. 

\rvhilight{
Here we do \emph{not} claim that $g_0$ or its capped counterpart $\bar{g}_L$ is \emph{the best} weight function for estimating truncated densities using generalized SM.  \eqref{eqn:Gamma_g1} and \eqref{eq.convergencebound} suggest that there should be other data-driven choices of $\boldg$ that yield better error bounds. However, $g_0$ should be an adequate choice without using any information on $q, p_\boldtheta$ and $V$, judging from our analysis. Finding a tractable data-driven $\boldg$ that minimizes the estimation error bound is an interesting future work (See Section 9 for more information). 
}

\subsection{Case Study: Truncated Density in a Unit Ball}
\label{sec:case_study}
In this section, we consider a case where $q$ has a parametric form, $V$ is a unit ball. Although this is a very specific setting, we can see how the relationship between $q$, $V$ and different choices of $L$ in $\bar{g}_L$ would influence the error bound. 
In what follows, we consider the capped distance weight $\bar{g}_{L}(\boldx) =\min\left\{\,L g_0(\boldx),\, 1\right\}$ which has been introduced in Corollary \ref{col.capped} where the distance weight $g_0$ is defined using the euclidean distance. 

Let us define $p_{\mathrm{dist}}(z)$ as the probability 
density of $Z=g_0(X)$ for $X\sim {q}$. 
For simplicity, we assume that there exist positive constants $b$ and $b'$ 
such that $b z^\beta\leq  p_{\mathrm{dist}}(z) \leq b' z^\beta$ holds for $0<z\leq c$, 
where $c$ and $\beta$ are  constants. 
Note that $\beta$ should be greater than $-1$ since the integral of $p_{\mathrm{dist}}(z)$ on the interval $(0,c)$ should be bounded. 

\begin{figure}
    \centering
\subfigure[$\beta = -.5$]{\includegraphics[width=.22\textwidth]{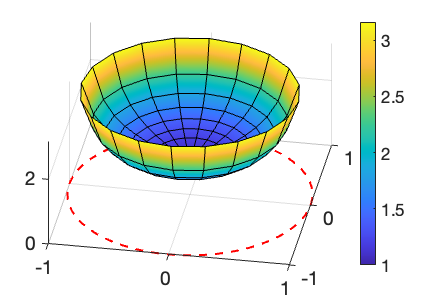}}
\subfigure[$\beta = 0$]{\includegraphics[width=.22\textwidth]{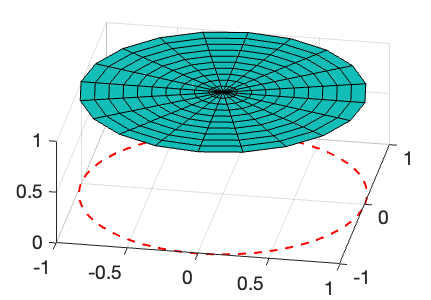}}
\subfigure[$\beta = 1$]{\includegraphics[width=.22\textwidth]{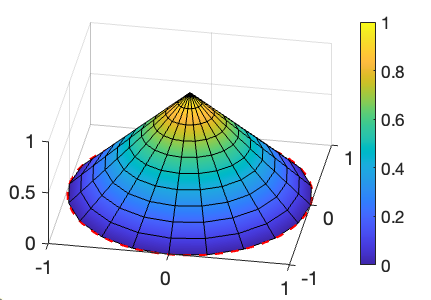}}
\subfigure[$\beta = 3$]{\includegraphics[width=.22\textwidth]{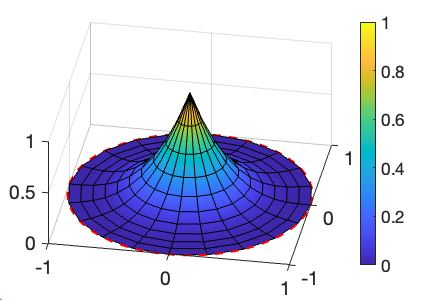}}\\
\subfigure[$\beta = -.5$]{\includegraphics[width=.22\textwidth]{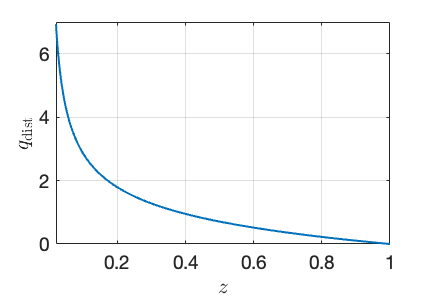}}
\subfigure[$\beta = 0$]{\includegraphics[width=.22\textwidth]{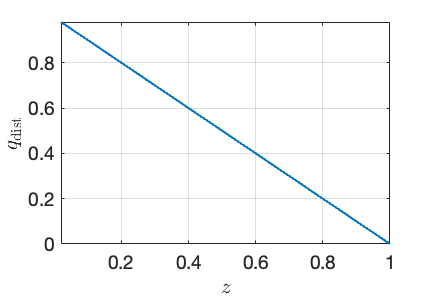}}
\subfigure[$\beta = 1$]{\includegraphics[width=.22\textwidth]{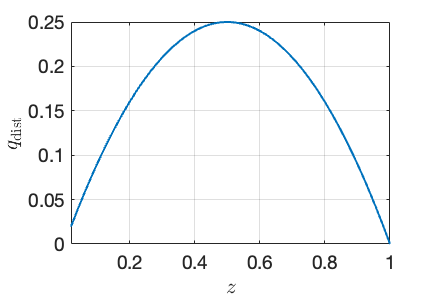}}
\subfigure[$\beta = 3$]{\includegraphics[width=.22\textwidth]{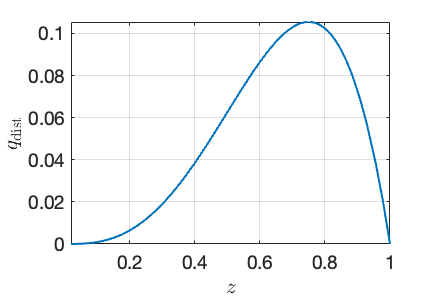}}
\caption{Top row: unnormalized $q_\beta(\boldx)$, where $\partial V$ in Example \ref{ex.unitball} is illustrated using a red dashed line. Bottom row: unnormalized $p_\mathrm{dist}(z)$  in Example \ref{ex.unitball}. 
}
\label{fig.q.pdist}
\end{figure}

\begin{exam}
\label{ex.unitball}
Let $V$ be the unit open $\ell^2$ ball in $\mathbbR^d$, i.e., $V=\{\boldx\in{\mathbb{R}^2}|\|\boldx\|<1\}$. 
We can consider a distribution $q_\beta(\boldx) \propto (1-\|\boldx\|)^\beta$ on $V$, where $\beta>-1$. 
 Then, we have $p_{\mathrm{dist}}(z) \propto (1-z)^{d-1}z^\beta,\, 0<z\leq 1$. 
 For a small $z$, we have $b z^\beta \leq p_{\mathrm{dist}}(z)\leq b'z^\beta$. 
\end{exam}
See Figure \ref{fig.q.pdist} for illustrations of unnormalized $q_\beta$ and $p_\mathrm{dist}(z;\beta)$. 
It can be seen that:
\begin{itemize}
    \item If $q(\boldx)$ converges to a positive constant as $\boldx\rightarrow\partial{V}$, 
     $\beta=0$.   
    \item If $q(\boldx)$ tends to zero, 
    as $\boldx\rightarrow\partial{V}$, $\beta>0$. 
    \item If $q(\boldx)$ tends to  infinity, 
    as $\boldx\rightarrow\partial{V}$, $\beta<0$.
\end{itemize}

For $\beta>-1$ and $L\ge \max\{1,1/c\}$, where $c$ is independent of $L$, we have the bounds for $\frac{\Gamma_{\bar{g}_L} }{C_{\bar{g}_L}}$:
\begin{align}
\label{eqn:lower-upper-Boundary}
C_{A}\left(1-\frac{C_{b',\beta}}{L^{\beta+1}}\right)^{1/4}\!\!+C_{B}c_0 L^{(1-\beta)/2}
\le
\frac{\Gamma_{\bar{g}_L} }{C_{\bar{g}_L}}
\le
C_{A}\left(1-\frac{c_{b,\beta}}{L^{\beta+1}}\right)^{1/4}\!\!+C_{B}c_1  L^{(3-\beta)/4}, 
\end{align}
where $c_0, c_1, c_{b,\beta}, C_{b',\beta}, C_A, C_B$ 
are positive constants independent of $L$.  
The first term of the lower and upper bounds is positive for $L\ge 1$. 
The detailed derivation of this bound is found in 
Appendix~\ref{Appendix:derivation_lower-upper-BoundaryCond}. Although we assume that $p_\mathrm{dist}$ takes a specific form in this example, the derivation mostly concerns the behavior of $p_{\mathrm{dist}}(z), z<c$. Thus a slight modification of \eqref{eqn:lower-upper-Boundary} should cover a different pair of $p_\mathrm{dist}$ and $q_\beta$ that has the same behavior near the boundary.

First, let us consider the condition $\beta>3$ in which case $q(\boldx)$ rapidly goes to zero
as $\boldx$ converges to a point on the boundary of $V$. 
The upper and lower bounds in \eqref{eqn:lower-upper-Boundary}
converge to $C_A$ as $L$ tends to infinity, meaning that 
a large $L$ does guarantee a reasonable accuracy. 
When $\beta > 3$, $q(\boldx)$ is almost zero around the boundary (see Figure \ref{fig.q.pdist}). By setting $L$ to a large value, TruncSM with $\bar{g}_L$ is essentially the classic SM which is well-suited for  non-truncated density estimation. 

On the other hand, when $q_\beta$ goes to zero slowly ($0<\beta\le1$) or converges to a constant at the boundary ($\beta \le 0$), 
a larger $L$ leads to a larger lower bound, which is undesirable. In particular, when $\beta <0$, increasing $L$ will increase both upper and lowerbound rapidly. This implies a smaller $L$ would yield a better performance when $\beta$ is small. 

Our analysis can also be  validated via Figure \ref{fig.q.pdist} and Theorem \ref{theorem:convergence_GSM}: When $\beta$ is large, $p_{\mathrm{dist}}$ is low around the boundary, thus a steep $\bar{g}_L$ (i.e., large $L$) near the boundary would not blow up  $\Gamma_\boldg$ (which depends on $\mathrm{E}_q [(\partial_k g_k)^4] \le L^4$). However, as we reduce $\beta$, $p_\mathrm{dist}$ takes higher values near the boundary. A steep $\bar{g}_L$ near the boundary would lead to a large $\Gamma_\boldg$, hence a larger estimation error.

  For the truncated distribution on the bounded domain such as the truncated Gaussian (or Gaussian mixture) model, 
  the probability density  $p_{\mathrm{dist}}(z)$ is greater than a positive constant near the boundary. The above analysis shows that 
  the capped distance function works efficiently
  for such truncated probability models. We will empirically validate this analysis in Section \ref{sec.weightfunc}.

\section{Computation of Distance Weight $g_0$}
\label{sec:computation}Comparing to other choices of weight functions, distance weight $g_0$ (and capped distance $\bar{g}_L$) have an important advantage: the computation of $g_0$ and its gradient can be efficiently carried out for a properly defined $V$. 

For example, if $V$ and $\partial{V}$ are expressed by
$V=\{ \boldx \in \mathbbR^d | u(\boldx) < 0\}$ and $\partial V= \{ \boldx \in \mathbbR^d | u(\boldx) = 0\}$ using a function $u:V\rightarrow\mathbbR$, 
evaluating $g_0(\boldx)$ can be turned into an optimization problem: 
$g_0(\boldx) = \min_{\boldz}\{\mathrm{dist}(\boldx,\boldz)\,|\,u(\boldz) = 0\}$.
In addition, if Euclidean distance is considered, the gradient of $g_0(\boldx)$ is simply given by 
$\nabla_\boldx g_0(\boldx)= (\boldx- \tilde{\boldx})/\|\boldx- \tilde{\boldx}\|$, 
where $\tilde{\boldx}$ is the minimizer of $\min_{\boldz\in \partial{V}} \|\boldx -\boldz\|$. 
We only need to evaluate $g_0$ and $\partial_k g_0$ \emph{exactly once} for all $\boldx \in X_q$ \emph{before} estimating $\hat{\boldtheta}$, since $g_0$ is model agnostic. This can be advantageous when $\bar{p}_\boldtheta$ is a sophisticated model and the optimization for $\boldtheta$ is time-consuming. 
In contrast, if one uses Monte Carlo methods (e.g. \citet{kannan1997random}) to approximate the normalizing constant $Z_V(\boldtheta)$ in the likelihood gradient, 
they are required to update the estimate of $Z_V(\boldtheta)$ throughout the entire gradient descent procedure. 
Approximating $Z_V(\boldtheta)$ using Markov Chain Monte Carlo (MCMC) \citep{robert2013monte} can also get less efficient both in terms of statistical accuracy and computation as dimensionality increases. 
Moreover, some algorithms designed for estimating the truncated multivariate Gaussian likelihood gradient, such as the ones proposed by \citet{daskalakis2018efficient,daskalakis19a}, do not require exhaustive sampling. 
They need to evaluate a membership oracle (if $\boldx \in V$ or not) for auxiliary samples freshly drawn at each iteration.  
\rvhilight{
Membership evaluation can be more efficient than computing $g_0(\boldx)$ for each $\boldx$. However, TruncSM only evaluates $g_0$ for  samples in the data set $X_q$ once, thus it is a fixed computation cost that does not grow with the number of gradient descent iterations. 

There may be a $U\subset V$, for all $\boldx \in U$, the corresponding $\tilde{\boldx}$ is not unique. If so,  $g_0$ will be non-differentiable on $U$. 
However, 
Lemma \ref{lem.2} states that the area in $V$ where $g_0$ is non-differentiable has a measure zero, thus $U$ also has a measure zero. Therefore, we do not need to worry about such a case in practice. 
}
In what follows, we show efficient analytical methods for computing the distance function defined over \emph{unit ball}, \emph{unit cube},  \emph{convex polytopes} and \emph{polygons}. 

\begin{itemize}
    \item For the Unit Ball, 
    $V = \{\boldx \in \mathbb{R}^d | \|\boldx\| < 1 \}$, the distance function and its gradient: 
    \[
        g_0(\boldx) = 1-\|\boldx\|, \nabla_\boldx g_0(\boldx) = \frac{-\boldx}{\|\boldx\|}.
    \]
    \item For the Unit Cube, 
    $V = \{\boldx \in \mathbb{R}^d | \|\boldx\|_\infty < 1 \}$, the distance function and its gradient:
    \[
        g_0(\boldx) = 1-\|\boldx\|_\infty, \ 
        \nabla_\boldx g_0(\boldx) = -\bolde_j,\ j = \argmax_k |x_k|. 
    \]
    \item         For the Convex Polytope $V=\{\boldx\in\mathbbR^d|\langle\bolda_t,\boldx\rangle+b_t<0, t=1,\ldots,T\}$, the distance function is given by 
\begin{align}
 \label{eq.poly.dist}
 g_0(\boldx) 
 = \min_{\boldz}\{\|\boldx-\boldz\|\,|\, \max_{t\in[T]}\{\langle\bolda_t,\boldz\rangle + b_t\} = 0\} 
 = \min_{t\in[T]}\frac{|\langle\bolda_t,\boldx\rangle + b_t|}{\|\bolda_t\|}
\end{align}
for $\boldx\in{V}$. The envelope theorem~\citep{paul02:_envel} yields $\nabla_{\boldx}g_0(\boldx)=-\bolda_{t^*}/\|\bolda_{t^*}\|$, 
where $t^*\in[T]$ is the minimizer of  \eqref{eq.poly.dist}. 
\citet{JOTA:Briec:1997} studied another representation of the minimum distance problem for the convex polyhedral using the extreme points. 
Note that if $V$ is not a convex set, \eqref{eq.poly.dist} cannot be used. 

\item For the Convex Polytope 
    $V=\{\boldx\in\mathbbR^d|\langle\bolda_t,\boldx\rangle+b_t<0, t=1,\ldots,T\}$, 
        let us compute the $\ell^1$-based distance function 
    $g_0(\boldx)=\min_{\boldz\in\partial{V}}\|\boldx-\boldz\|_1$ for $\boldx\in{V}$. 
It holds that 
$g_0(\boldx) =  \max\{|\alpha|\,|\,\boldx+\alpha \bolde_i \in {V},\ \forall{i}\}$
since $V$ is convex. 
The condition $\boldx+\alpha \bolde_i \in {V},\ \forall{i}$ is expressed by 
$\alpha \langle\bolda_t,\bolde_i\rangle\leq -\langle\bolda_t,\boldx\rangle-b_t$ 
for all $t$ and $i$. 
Hence, we have 
 \begin{align}
 \label{eq.poly.L1dist}
  g_0(\boldx) 
  = \min_{i,t\,\text{s.t.}\,\langle\bolda_t,\bolde_i\rangle\neq0} \left|\frac{\langle\bolda_t,\boldx\rangle+b_t}{\langle\bolda_t,\bolde_i\rangle}\right|
  = \min_{t\in[T]}\frac{|\langle\bolda_t,\boldx\rangle+b_t|}{\|\bolda_t\|_\infty}. 
 \end{align}
 The envelope theorem~\citep{paul02:_envel} yields $\nabla_{\boldx}g_0(\boldx)=-\bolda_{t^*}/\|\bolda_{t^*}\|_{\infty}$ for $\boldx\in{V}$, 
 where $t^*\in[T]$ is the minimizer of the last minimization in~\eqref{eq.poly.L1dist}.  
 
 \item For the Polygon $V$ in $\mathbbR^2$ surrounded by the points 
$\boldp_1, \boldp_2,\ldots,\boldp_T\in\mathbb{R}^2$, the boundary is given by 
$\partial{V}=\cup_{t=1}^{T}\{\alpha\boldp_t+(1-\alpha)\boldp_{t+1}|\alpha\in[0,1]\}$, where 
$\boldp_{T+1}=\boldp_1$. 
Hence we have 
\[g_0(\boldx)=\min_{t\in[T]}\min_{\alpha:0\leq \alpha\leq 1}\|\boldx-\alpha\boldp_t-(1-\alpha)\boldp_{t+1}\|.\]
The minimizer of the inner optimization is
$\alpha_t=\min\{1, \max\{0, \frac{\langle\boldp_{t}-\boldp_{t+1}, \boldx-\boldp_{t+1}\rangle}{\|\boldp_{t}-\boldp_{t+1}\|^2}\}\}$. 
Hence, we obtain $g_0(\boldx)=\min_{t\in[T]}\|\boldx-\alpha_t\boldp_t-(1-\alpha_t)\boldp_{t+1}\|$, and
$\nabla_{\boldx}g_0(\boldx)=\bold{n}/\|\bold{n}\|$, where $\bold{n}=\boldx-\alpha_{t^*}\boldp_{t^*}-(1-\alpha_{t^*})\boldp_{t^*+1}$ 
and where $t^*\in[T]$ is the minimizer of $g_0(\boldx)$. 

\end{itemize}

In these cases, the computation of $g_0$ can be done in polynomial time with respect to $d$ (if $T$ in convex polytope is a polynomial function of $d$).

\section{Numerical and Real-world Data Analysis}
\label{sec:Experiments}
To demonstrate the efficacy of TruncSM empirically, we conduct a wide range of experiments. In this section, we use the Euclidean metric for $g_0$ and $\bar{g}_L$. The code/data sets to reproduce our experiments are available at 
\url{https://github.com/anewgithubname/Truncated-Score-Matching}. 

\subsection{Illustrative Example and Computation Time}
\label{sec.toy.1}
\begin{figure}
    \centering
    \subfigure[Estimation of truncated Gaussian mixture centers]{    \includegraphics[width =.49\textwidth]{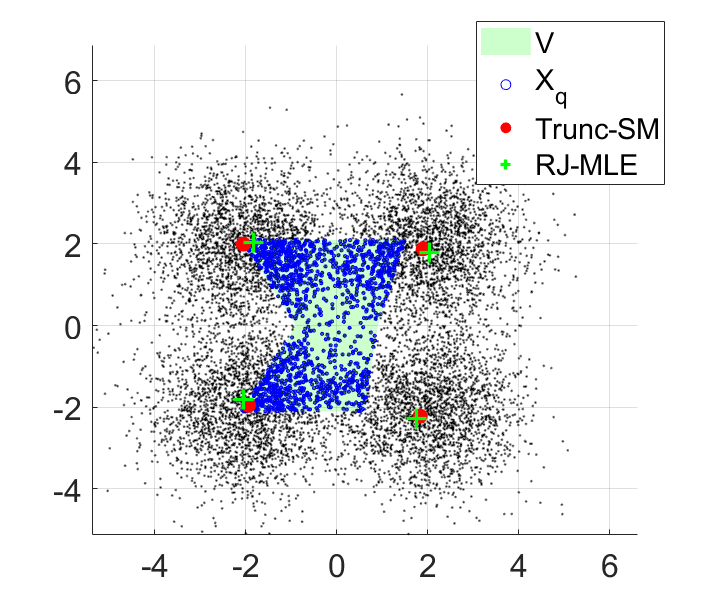}
    \label{fig:truncGM.illus}
    }
    \subfigure[Computational cost and estimation accuracy comparison for TruncSM (red) and RJ-MLE (blue)]{    \includegraphics[width =.46\textwidth]{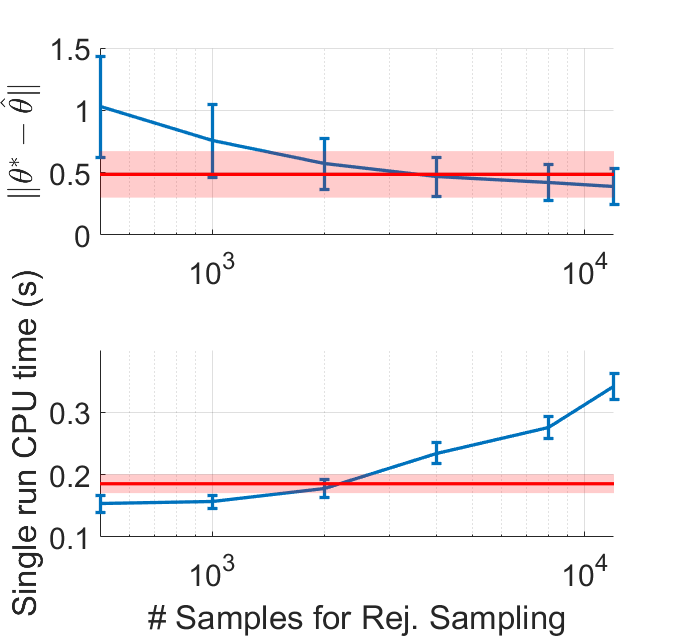}
    \label{fig:truncGM.comp}
    }
    \caption{Gaussian mixture centers truncated by a polygon}
\end{figure}

In the first experiment, we show an illustrative example comparing TruncSM and Rejection Sampling MLE (RJ-MLE), as well as their computational times. 
\rvhilight{
RJ-MLE uses rejection sampling to approximate the intractable normalizing term $Z_V(\boldtheta)$ and perform maximum likelihood estimation. It can be seen as an example of Monte Carlo MLE \citep{Geyer1994}.}
Samples are generated from a Gaussian mixture on $\mathbbR^2$, with centers at 
\[\boldsymbol{\mu}_1 =[2,2], ~~ \boldsymbol{\mu}_2=[-2,2], ~~ \boldsymbol{\mu}_3=[-2,-2],
~~\boldsymbol{\mu}_4=[2,-2], \]  and standard deviations all set to 1. 
The pre-truncated data set can be seen in Figure \ref{fig:truncGM.illus} as black dots.  
To create a truncated data set, we limit our observation window to be a green polygon region in the middle, 
thus only samples inside the green polygon (blue points) can be observed. 
The task is to find all four centers of the mixture model using \emph{only blue points}. 
We generate 10,000 samples and only 1417 of which within the truncation boundary can be used for parameter estimation.
Our unnormalized density model is a Gaussian mixture model with four components (parametrized by $\boldtheta_{1},\ldots,\boldtheta_{4}$) and the unit variance-covariance matrix: 
$\bar{p}_{\boldtheta_{1},\ldots,\boldtheta_{4}}(\boldx) = \sum_{i=1}^{4}\mathcal{N}_\boldx(\boldtheta_i,\boldI).$

As the polygon boundary $\partial{V}$ of the non-convex domain $V$ in $\mathbbR^2$ consists of line segments, 
the analytical algorithm in Section~\ref{sec:computation}
is available to compute $g_0$ and $\nabla_{\boldx}g_0$ efficiently. 
We compare with RJ-MLE which uses rejection sampling to approximate the normalizing constant $Z_V(\boldtheta):= \int_V \bar{p}_{\boldtheta_{1},\ldots,\boldtheta_{4}}(\boldx) \dx$. In this experiment, 500,000 particles are used to approximate $Z_V(\boldtheta)$ and they are drawn from a bivariate uniform distribution with density $\mathcal{U}_{x_1}(-2,2) \cdot \mathcal{U}_{x_2}(-2,2)$. 
The estimated mixture component centers are plotted as green crosses (RJ-MLE) and red dots (TruncSM) in Figure \ref{fig:truncGM.illus}. It can be seen that both methods give estimates close to the true mixture centers. However, the computation time for TruncSM and RJ-MLE are 0.35 seconds and 3.35 seconds respectively\footnote{For TruncSM, we include the calculation time for both $g_0$ and $\partial_k g_k$. Same below.} \footnote{Our experiments are run on a workstation with a AMD Ryzen 1700 CPU with 32GB memory.}. 

It is worth noting that our particle distribution is carefully chosen so that it tightly covers the truncation domain. It ensures the best rejection sampling performance: Over-coverage would increase the computational cost and insufficient coverage would lower the approximation accuracy of $Z_V(\boldtheta)$. 
While in a lower dimensional space, the visualization helps, in a higher dimensional space, it is  unclear how to come up with a good rejection sampling distribution.

To further investigate the computation time between TruncSM and RJ-MLE, we study the estimation accuracy and computation time (with standard deviation) against the number of particles used for rejection sampling by RJ-MLE. We optimize both objective functions using MATLAB's \verb|fminunc| function with default settings. From Figure \ref{fig:truncGM.comp}, we can see that when using a large number of particles to approximate $Z_V(\boldtheta)$, RJ-MLE can indeed achieve a slightly better performance than TruncSM. However, such a slight improvement of estimation accuracy  comes with a significant penalty in computation cost even with an optimal choice of the 
rejection sampling distribution. 

\subsection{Capped Weight Function}
\label{sec.weightfunc}
\begin{figure}
    \centering
    \subfigure[]{\includegraphics[width = .7\textwidth]{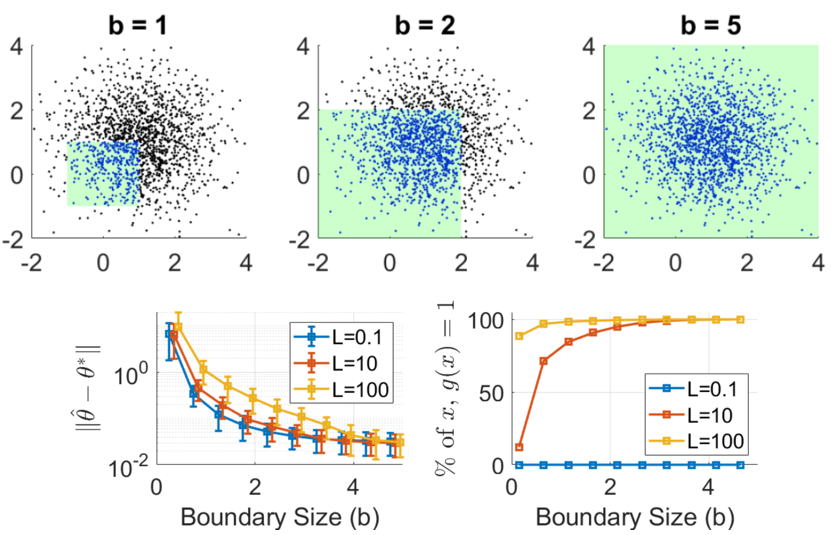}
    \label{fig.capped.joined}
    }
    \subfigure[]{\includegraphics[width = .7\textwidth]{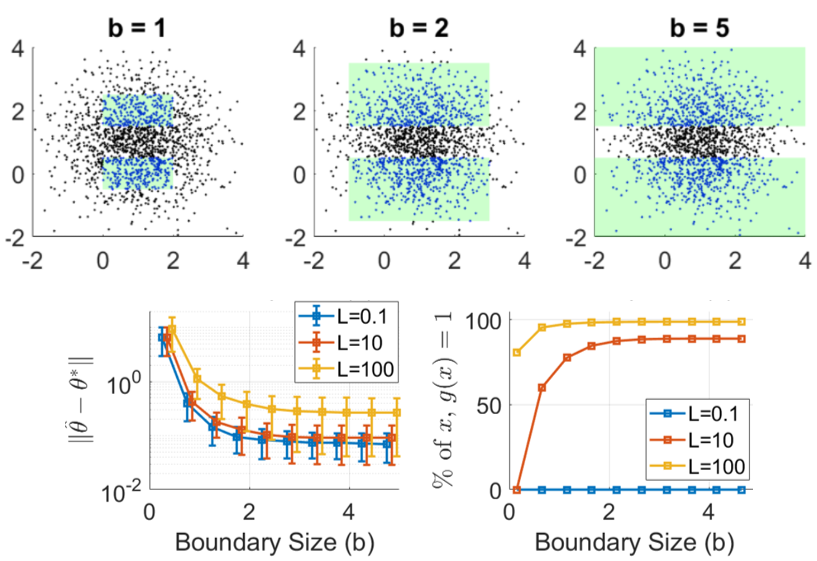}
    \label{fig.capped.disjoined}
    }
    \caption{The truncated data sets, the estimation performance and the percentage of data points whose $\bar{g}_L$ are capped as $V$ enlarges. $b$ is the scaling multiplier of polygon vertices.
    It can be seen that as the area of $V$ grows, $\bar{g}_L(\boldx)$ becomes capped for more and more samples in our data set (the bottom right plots in both (a) and (b)). This is expected as the boundary stretches, fewer and fewer points are adjacent to the boundary. 
    }
    \label{fig:boundary}
\end{figure}
Now we investigate the performance of a capped weight function: 
$\bar{g}_L := \min(1, L\cdot g_0(\boldx))$. It can be seen that when $L$ is small, 
$\bar{g}_L$ will never be capped, thus it is equivalent to $g_0$ after multiplying a constant, which does not affect the estimator. 
In this experiment, two different truncation domains are used: a single rectangle and two disjoint rectangles. 
1600 samples are drawn from a normal distribution $\mathcal{N}([1, 1], \boldI)$. 
We monitor the performance of TruncSM using $\bar{g}_L$ as the weight function as $V$ grows in size. 
We measure the growth of $V$ using a scaling factor $b$ 
(see Appendix~\ref{appendix:boundary_size} %
on how polygons are re-scaled). 
As shown in Remark~\ref{rem:non-connected}, the true parameter is identifiable by TruncSM even when the domain consists of two disjoint rectangles. 

Figure \ref{fig:boundary} illustrates the truncated data sets as $V$ grows, the estimation performance and the percentage of data points whose $\bar{g}_L$ are capped. 

In Figure \ref{fig.capped.joined}, 
we can observe that the performance gap between $L = 0.1$, $L = 10$ and $L = 100$ widens then shrinks: If the truncation domain is small, not many samples are included in the truncation domain. Thus algorithms with all choices of $L$ would suffer. However, as the truncation boundary grows, the difference starts to show: TruncSM with a smaller $L$ has a better performance as we analyzed in Section~\ref{sec:case_study}.  %
As $V$ grows beyond a certain point, the data set essentially becomes non-truncated, and TruncSM using large $L$ reduces to a classic SM since $\bar{g}_L$ are always capped at 1. At this time, TruncSM with all choices of $L
$
converges to the same level of performance. 
Figure \ref{fig.capped.disjoined} shows a similar story but estimation error with different $L$ converge to different levels as $V$ enlarges: 
$V$ never covers the center of our data set thus our data sets are always truncated. 
Again TruncSM with a smaller $L$ gives a better performance as we analyzed in Section \ref{sec:case_study}.

\subsection{2008 Chicago Crime data set}
\label{sec.chi}
\begin{figure}
 \centering 
  \includegraphics[width=0.45\textwidth]{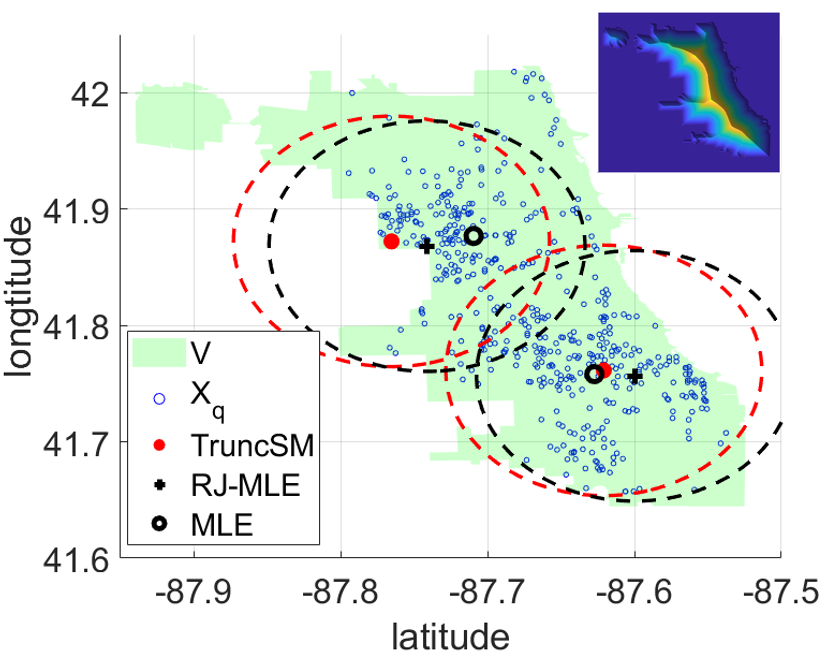}
  \caption{Chicago Crime data set, whose truncation boundary is a polygon. Blue circles are homicide locations. 
  $g_0$ is visualized at the upper right corner. The estimated component centers of TruncSM, RJ-MLE and MLE are plotted.}
  \label{fig.chi}
\end{figure}
We also test the performance of TruncSM on a real-world truncated density estimation problem: Analyzing the crime occurrences in Chicago. The data set contains locations of homicides that happened in Chicago during 2008. We fit a Gaussian mixture model with two components on this data set. The standard deviations of two components are fixed to the same value, which is
roughly the half of the ``width'' of the city.  

\rvhilight{In this experiment, we compare TruncSM with vanilla MLE using the non-truncated density model (MLE for short) and RJ-MLE using the truncated density model.
In this data set, Chicago city boundary is expressed via a polygon in $\mathbb{R}^2$, so the distance weight $g_0$ is calculated using the analytical solution given in Section \ref{sec:computation}. }

The estimated means of two components are plotted on Figure \ref{fig.chi}. 
The estimated 95\% confidence region is plotted for TruncSM and RJ-MLE as red and black dotted circles respectively. It can be seen that TruncSM, MLE and RJ-MLE all picked centers at the north and south side of the city. However, MLE picked a northern location inside of the city while TruncSM and RJ-MLE picked a location right next to the western border of Chicago.  

In this case, TruncSM and RJ-MLE tend to put observed crimes on the decaying slope of a Gaussian density which would better explain the declining
rate of crime from the west to the east. MLE, unaware of the truncation, puts the Gaussian center in the middle of the city, while clearly the crimes happen more rarely in the east.  

Although all estimators tested in this section solve non-convex optimization problems, in the vast majority of runs with different initializations, 
we observe only very minor changes in terms of estimated Gaussian centers between different runs. See Appendix~\ref{appendix:Chicago} for more discussion on this. 

\subsection{Outlier Over-trimming Compensation}
Removing outliers is an important data preprocessing step. 
If we know the percentage of outliers in our data set, 
we can adopt methods such as One-class Support Vector Machines (OSVM) \citep{scholkopf1999support} to remove them. 
However, it is often impossible to determine the percentage of outliers a priori and 
setting the outlier percentage aggressively may result in inliers also being removed from the data set. 

Consider outlier trimming using OSVM. The method outputs a ``decision function'' $u(\boldx) := \sum_j \hat{\alpha}_j \phi_j(\boldx)$ which defines a domain $V := \{\boldx \in \mathbb{R}^d | u(\boldx) >0\}$, where $\hat{\alpha}_j$ is the parameter obtained from OSVM procedure and $\phi_j$ is a kernel function. In this experiment, we use Gaussian kernel, defined as $\phi_j(\boldx) := \exp(-\|\boldx_j- \boldx\|^2/2\sigma^2)$, where $j$ is the $j$-th datapoint in the contaminated data set. 
OSVM chooses a $V$ such that a certain proportion (e.g. 80\%) of our data set is included in $V$. 
We discard any data point that is not in $V$ as ``outliers''. 
However, if the specified outlier percentage is larger than it actually is, we will trim inliers from our data set too, and our data set become a truncated data set. Estimation without considering the truncation boundary would lead to biased estimates. 

\begin{figure}[t]
    \centering
    \includegraphics[width=.47\textwidth]{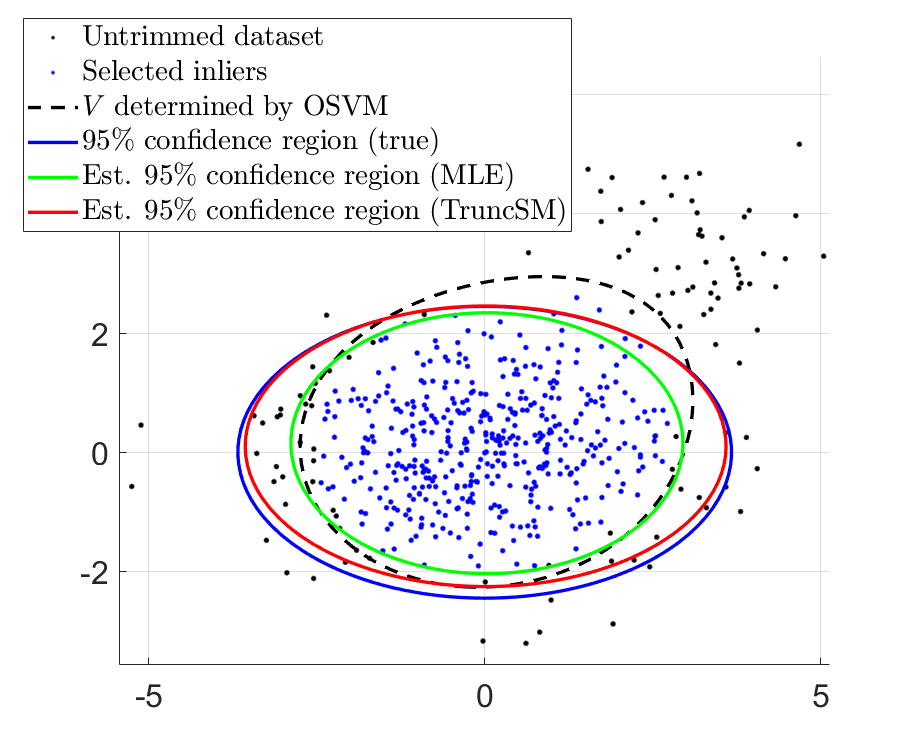}
    \includegraphics[width=.47\textwidth]{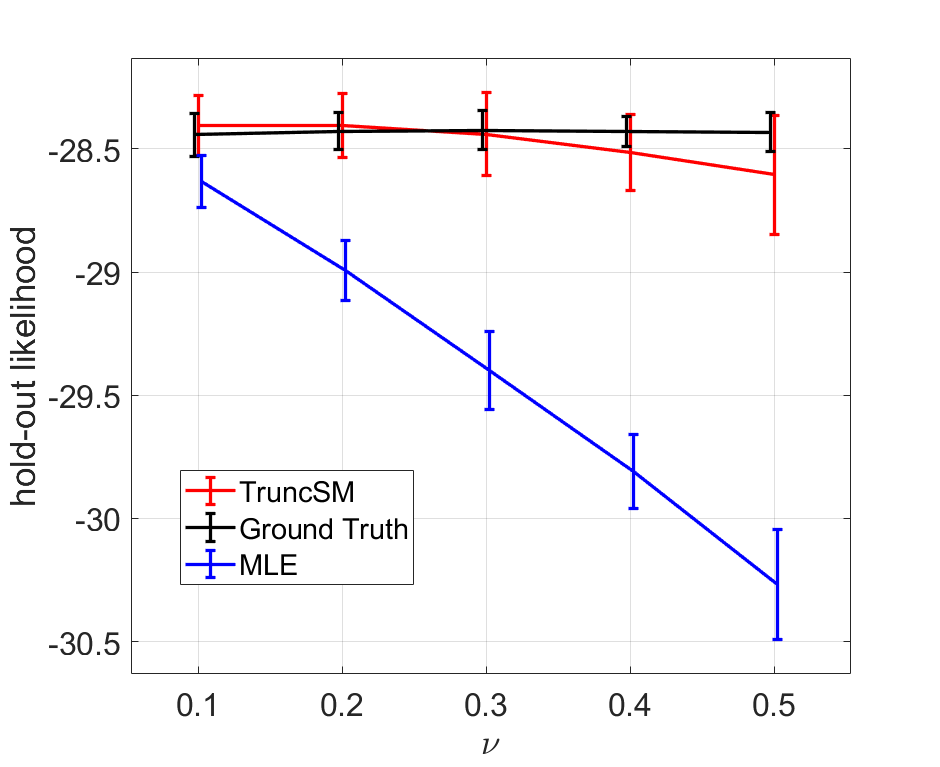}
    \includegraphics[width=.98\textwidth]{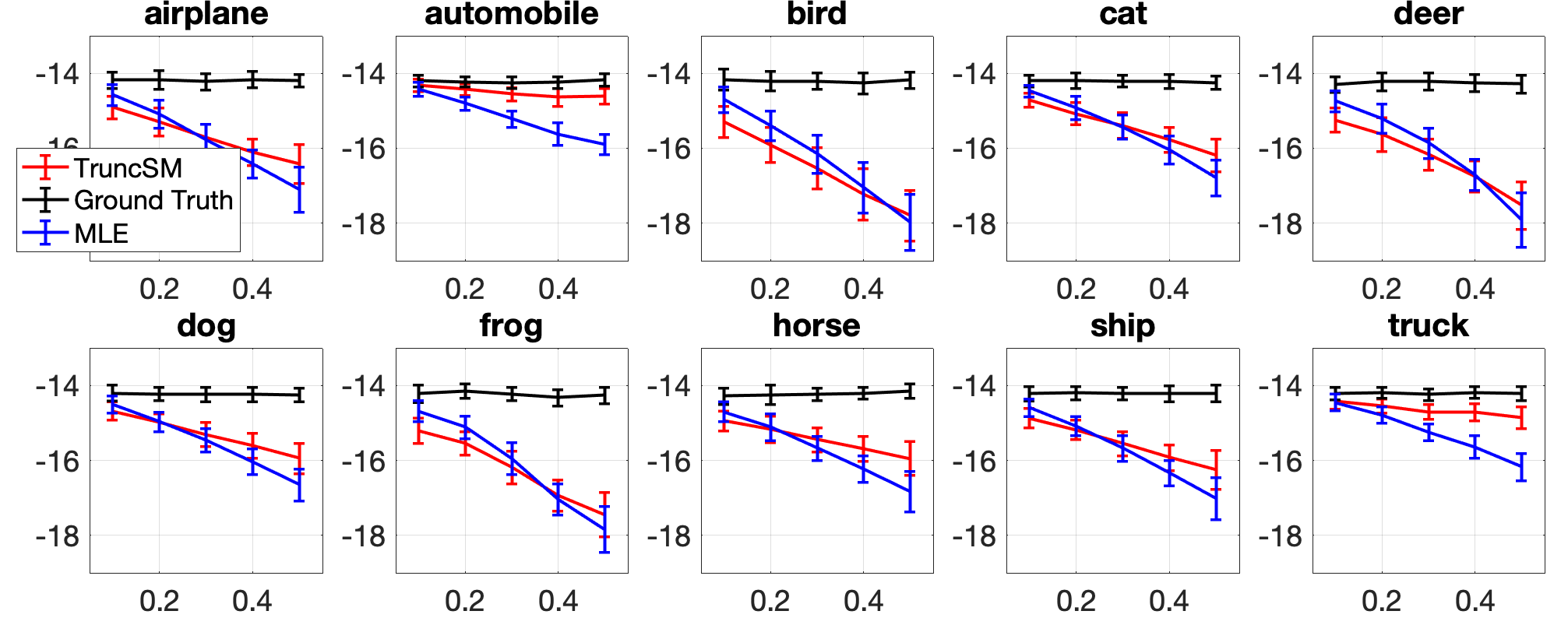}   
    \caption{Top left: MLE underestimates the 95\% confidence region due to the truncation. The confidence region estimated by TruncSM is much closer to the true confidence region.
    Top right, MLE getting less and less accurate as the percentage of truncated samples (controlled by $\nu$) increases while TruncSM maintaining a good accuracy. 
    Bottom two rows: TruncSM achieves a comparable or better performance than vanilla MLE as the percentage of truncation increases on CIFAR-10 data set.
    }
    \label{fig:onesvm1}
\end{figure}

This is demonstrated using the following simulated experiment. We sample 
\[
\boldx_{\text{inlier}} \sim \mathcal{N}\left(\boldzero, 
\begin{bmatrix}
1.5^2&0\\
0&1
\end{bmatrix}
\right), \qquad
\boldx_{\text{outlier}} \sim \mathcal{N}\left( \begin{bmatrix}
3.5\\
3
\end{bmatrix}, 
\begin{bmatrix}
.8^2&0\\
0&.8^2
\end{bmatrix}
\right).
\] In total, 500 inlier samples and 50 outlier samples are drawn. The outlier percentage ($\nu$) in OSVM is set to 20\%. 
The data set (black dots), the selected inliers (blue dots) and the truncation domain $V$ given by OSVM is visualized in the top left plot in Figure \ref{fig:onesvm1}. In this experiment, we allow MATLAB to automatically determine the kernel bandwidth $\sigma$ according its predefined protocol. On one hand, it can be seen that OSVM does separate the inliers from outliers using a boundary function $u(\boldx)$. On the other hand, some inlier samples are also removed due to the aggressive setting of the outlier proportion.

We use the trimmed data set to estimate a truncated multivariate normal distribution $p_{\boldmu, \boldSigma}(\boldx) \propto \mathcal{N}(\boldmu, \boldSigma)$, where $\boldSigma$ is restricted to be a diagonal matrix. 
In this experiment, TruncSM uses the distance weight $g_0$. $g_0$ and $\nabla_\boldx g_0$ are computed numerically using the constrained optimization described in Section \ref{sec:computation}. We compare TruncSM with vanilla MLE which does not use any truncation information. 
The true and estimated 95\% confidence regions are visualized by ellipses in Figure \ref{fig:onesvm1}.  The vanilla MLE underestimates the variance of the inlier distribution, while TruncSM accurately recovers the 95\% confidence region using the estimated inlier density function. We then perform the same experiment in a much higher dimensional space: The inlier distribution is a 20 dimensional standard normal distribution while the outlier distribution is a 20 dimensional normal distribution with mean $\boldone$ and unit variance. 
Note in this experiment, the truncation boundary (induced by a kernel function) is highly irregular. 

In the top right plot of Figure \ref{fig:onesvm1}, we plot the hold-out likelihood versus different settings of outlier proportion ($\nu$) in OSVM. The likelihood is computed using both vanilla MLE and TruncSM solutions. 
As we can see, the more aggressive our outlier trimming is, the worse the vanilla MLE performs. On the other hand, TruncSM only drops very slightly in performance as more and more inlier data points are truncated. 

We now perform an experiment on a real-world data set CIFAR-10 which contains 10 different classes of 32 by 32 images.
To speed up the computation, we reduced the dimension of the data set to 10 using PCA. 
For each class, we artificially add 10\% outliers that are drawn from a normal distribution with mean $\boldone$ and unit variance.
We use both TruncSM and vanilla MLE to estimate a multivariate Normal distribution for each class in the reduced 10 dimensional space and plot the hold-out likelihood of each method. 
The hold-out likelihood is plotted at the bottom two rows of Figure \ref{fig:onesvm1}. For a large $\nu$, we find that TruncSM performs better than MLE. The holdout likelihood shows that the estimation bias induced by the trimmed data can be corrected by TruncSM. Although the advantage of TruncSM is smaller on this data set, we argue that this is a very challenging problem: The data set is hardly Gaussian so the model assumption used by TruncSM is wrong. Nonetheless, in some cases (automobile and truck), TruncSM  significantly outperforms MLE.

\section{Conclusions and Future Works}
We propose an estimator for truncated statistical models with complex  truncation boundaries based on generalized SM. 
The proposed method uses the shortest distance (or capped distance) from a data point to the truncation boundary as the weight function. 
Such a choice of weight function naturally arises from minimizing Stein Discrepancy and lowering the estimation error upper bound. 
The proposed weight function is also computationally favorable for high dimensional truncation domains. 
Experiments on synthetic data and the Chicago crime data set show promising results. The proposed estimator was later applied to outlier trimming bias correction. 

Although the proposed method achieves promising results in truncated density estimation, there are interesting open questions:
\begin{itemize}

    \item \rvhilight{How to choose an optimal weight function $\boldg$ when $p_\boldtheta, q$ and $V$ are given? As we have seen from \eqref{eq.convergencebound} in Theorem \ref{theorem:convergence_GSM}, the statistical estimation error depends on the ratio $\frac{\Gamma_\boldg}{C_\boldg}$ and both $\Gamma_\boldg$ and $C_\boldg$ depends on the weight function $\boldg$, the density model $p_\boldtheta$,  the data density $q$ and the truncation boundary $V$. A natural idea is to choose a $\boldg$ that minimizes $\frac{\Gamma_\boldg}{C_\boldg}$. Can we find an efficient numerical procedure for such a minimization? }

    \item A related question is, how to choose an appropriate  distance metric in $g_0$? For example,  $\ell^1$ and $\ell^2$ distances are both valid choices for $g_0$. Again, 
    Theorem \ref{theorem:convergence_GSM} suggests that, in terms of upperbounding the statistical estimation error, the answer depends on $q$, $p_\boldtheta$ as well as $V$. However, how would the geometry of $V$, $q$ and $p_\boldtheta$
    affect the choice of the distance metric? 
    See Appendix \ref{appendix:ell1_ell2} for an empirical 
    comparison between $\ell^1$ and $\ell^2$ distances. 
    
    \item \rvhilight{The computation of $g_0$ for a generic $V$ is not trivial. In particular, when the dimensionality of data is large, computing $g_0$ can become computationally infeasible. Finding efficient ways of evaluating or approximately evaluating $g_0$ may help us extend the usage of TruncSM estimator to higher dimensional data sets. }

    \item In many applications, our data set is automatically filtered by some algorithm. Outlier trimming by OSVM is just one example.  Suppose we have mixed images of cats and dogs and a binary classifier, we can easily identify the high confidence region of each class. However, the classification step would also create artificial truncation boundaries around the distributions of individual classes. Can we use TruncSM to accurately reconstruct the true underlying distribution using these pre-classified images?  

\end{itemize}

\acks{
We thank two anonymous reviewers for their insightful feedbacks. 
This work was supported by Japan Society for the Promotion of Science under KAKENHI Grant Number 17H00764, 19H04071, and 20H00576. Daniel J. Williams was supported by a PhD studentship from the EPSRC Centre for Doctoral Training in Computational Statistics and Data Science (COMPASS). }

\renewcommand{\theequation}{\Alph{section}.\arabic{equation}}

\renewcommand{\themytheorem}{\Alph{section}.\arabic{mytheorem}}

\newpage

\begin{appendices}

\section{Generalized Score Matching}
\label{appenidx:GeneralizedScoreMatching}
Assume that $p(\boldx), q(\boldx)$ and $\boldg(\boldx)$ take strictly positive values on the domain $V\subset\mathbbR^d$. 
Suppose that generalized SM objective function with weight $\boldg$ is equal to zero, i.e., 
\begin{align}
 \mathbbE_{q}[\|\boldg^{1/2}(\boldx) \circ \nabla_\boldx \log p(\boldx)  -  \boldg^{1/2}(\boldx) \circ \nabla_\boldx \log q(\boldx)  \|^2] =0. 
\end{align}
Then, we have $\nabla_\boldx (\log p(\boldx) - \log q(\boldx))=\bold{0}$, meaning that
 $p(\boldx)= C q(\boldx)$ on (the connected domain) $V$ with a positive constant $C$. Since both $p$ and $q$
 are the probability densities on $V$, we have $C=1$. 

\section{Proof of Lemma \ref{lem.2}}
\label{appendix:lemma2}
\begin{proof}
	First, we show that $g_0$ is Lipschitz continuous with respect to the metric $\mathrm{dist}(\cdot, \cdot)$.
	\begin{align*}
		\forall \boldx_a, \boldx_b \in V,
		g_0(\boldx_a) - g_0(\boldx_b) 
		&=   \left(\min_{\boldx'\in \partial V}\max_{\boldx''\in \partial V} \mathrm{dist}(\boldx_a, \boldx') - \mathrm{dist}(\boldx_b,  \boldx'')\right)\notag\\
		&\le  
		\left(\max_{\boldx''\in \partial V} \mathrm{dist}(\boldx_a , \boldx'') - \mathrm{dist}(\boldx_b , \boldx'')\right) 
		\le \mathrm{dist}(\boldx_a, \boldx_b). 
	\end{align*}
	The last inequality is due to the triangle inequality. 
	Likewise, $g_0(\boldx_b)-g_0(\boldx_a)$ is also bounded above by 
	$\mathrm{dist}(\boldx_a, \boldx_b)$. Therefore $g_0$ is a Lipschitz function with Lipschitz constant 1. 
	Rademacher's theorem~\citep{evans92:_measur_theor_fine_proper_funct} 
	asserts that a Lipschitz continuous function is differentiable at every point in a Lipschitz domain outside a set of \emph{measure zero}. Therefore,
	we can construct 
	\begin{align}
		\label{eq.weakdiff_g0}
		D_k g_0(\boldx) := \begin{cases}
			\partial_k g_0(\boldx), & \text{if } g_0 \text{ is differentiable},\\
			\text{arbitary constant}, & \text{otherwise}
		\end{cases}
	\end{align}
	We can check that $D_k g_0(\boldx)$ is a valid weak derivative (Definition 7.1.3., \citep{atkinson2005theoretical}): 
	\begin{align*}
		\int_V g_0(\boldx) \partial_k \phi(\boldx) \dx = - \int_V D_k g_0(\boldx) \phi(\boldx) \dx,
	\end{align*} 
	where $\phi$ is any $m$-times differentiable function on a compact support on V.
	The equality holds due to the classic integration by parts formula for continuous differentiable functions. $g_0$ is only non-differentiable over a zero set, so the arbitary constant we set in \eqref{eq.weakdiff_g0} does not affect the outcome of the integration. 
	Since $V$ is a bounded domain and $g_0$ is 1-Lipschitz, $g_0$ and $D_k g_0$ are both bounded in terms of the $\|\cdot\|_{L^2(V)}$ norm. 
	Therefore, $g_0$ is weakly differentiable and in a Sobolev-Hilbert space. 
\end{proof}

\section{Proof of Theorem~\ref{theorem:general_objfunction}}
\label{appendix:Proof_Trunc-SM-tractable}
\begin{proof}
  In this proof, we apply Theorem \ref{thm.greenI} to derive a tractable expression for $M(\boldtheta)$.
  Let us consider the second term of $M(\boldtheta)$. %
  As $q(\boldx)$ and $g_0(\boldx)\partial_k \log{p_\theta}(\boldx)$ are functions in the Sobolev-Hilbert space
  of the first order, the direct application of Theorem \ref{thm.greenI} leads to 
  \begin{align*}
  &\phantom{=} 
  \sum_{k=1}^{d} \int_{V}g_k(\boldx)[{\partial}_k\log p_{\theta}(\boldx)] [{\partial}_k\log q(\boldx)] q(\boldx) \mathrm{d}\boldx \\ 
  &=
   \sum_{k=1}^{d}
   \int_{V} g_k(\boldx) [{\partial}_k\log p_{\theta}(\boldx)] [{\partial}_k q(\boldx)] \mathrm{d}\boldx  \\ 
   &= \sum_{k=1}^{d}\left\{ 
   \int_{\partial V} g_k(\boldx) [{\partial}_k\log p_{\theta}(\boldx)] q(\boldx) \nu_k(\boldx) \mathrm{d}s 
   -\int_{V}\partial_k[g_k(\boldx) {\partial}_k \log p_{\theta}(\boldx)] q(\boldx) \mathrm{d}\boldx\right\}\\
  &=
   -\sum_{k=1}^{d}\int_{V}\partial_k[g_k(\boldx) {\partial}_k \log p_{\theta}(\boldx)] q(\boldx) \mathrm{d}\boldx. 
  \end{align*}
  The second equality is ensured by Theorem~\ref{thm.greenI} and the third equality holds from the boundary condition imposed in the theorem. 
  \end{proof}
\section{Proofs of Theorem~\ref{thm.stein.dis} and Corollary~\ref{col.capped} in Section~\ref{subsec:TruncatedScoreMatching_MSD}}
\subsection{Proof of Lemma \ref{lem.minmax.SM}}
\label{sec.minmax.SM}
\begin{align*}
     &\max_{g_k \in \mathrm{Lip}^L_0(V)} \mathbbE_q \left[ \sum_k (\partial_k \ell_\boldtheta - \partial_k \logq)^2 g_k\right] 
    = L \cdot \mathbbE_q \left[ \sum_k (\partial_k \ell_\boldtheta - \partial_k \logq)^2 g_0\right]= L \cdot \mathrm{FH}_{g_0}(q,p_\boldtheta),
\end{align*}
The first equality holds since $\forall k, (\partial_k \ell_\boldtheta - \partial_k \logq)^2 \ge 0$. Since for all $g_k \in \mathrm{Lip}_0^L(V)$, $g_k(\boldz) = 0,\forall \boldz \in \partial V$, 
  \begin{align}
    \label{eq.lem2.2}
    g_k(\boldx) = g_k(\boldx) - g_k(\boldx') \le Ld(\boldx ,\boldx'), \forall \boldx'\in \partial V \implies g_k(\boldx) \le L \cdot \min_{\boldz\in \partial V} d(\boldx, \boldz) = Lg_0(\boldx),
  \end{align}
Hence the second equality. 

\subsection{Proofs of Theorem~\ref{thm.stein.dis}}
\label{appendix:TruncSM_MSD}
\begin{proof}
As we stated in the proof of v Lemma \ref{lem.2}, due to Rademacher's theorem~\citep{evans92:_measur_theor_fine_proper_funct}, any Lipschitz function defined on a Lipschitz domain is in $H^1(V)$, hence $g_k \in H^1(V)$ and $\forall k, f_k  \in H^1(V)$. Using the fact that $f_k, q,\ell_\boldtheta \in H^1(V), f_k(\boldx) = 0,  \forall \boldx \in \partial V$, we can verify that $\boldf$ is in a Stein class of $q$ by applying the same integration by parts techniques used in Section \ref{appendix:Proof_Trunc-SM-tractable}.
Therefore, we can write the Stein discrepancy between $p_\boldtheta$ and $q$ as: 
\begin{align}
\label{eq.maximum.rewrite}
    \max_{\boldf\in \mathcal{F}} \mathbbE_q \left[T_{p_\boldtheta} \boldf\right] = \max_{\boldf\in \mathcal{F}} \mathbbE_q \left[T_{p_\boldtheta} \boldf - T_q \boldf  \right].
\end{align}
Now we optimize \eqref{eq.maximum.rewrite} analytically using Theorem 2 in \citet{BarpMinimumStein2019} 
\begin{align*}
    \max_{\boldf\in \mathcal{F}} \mathbbE_q \left[T_{p_\boldtheta} \boldf - T_q \boldf  \right] &= \max_{g_k \in \mathrm{Lip}^L_0(V)}\max_{\boldh} \mathbbE_q \left[ \sum_k (\partial_k \ell_\boldtheta - \partial_k \logq) \cdot g_k^{\frac{1}{2}} h_k  \right] \\
    &= \max_{g_k \in \mathrm{Lip}^L_0(V)} \sqrt{\mathbbE_q \left[ \sum_k (\partial_k \ell_\boldtheta - \partial_k \logq)^2 g_k\right]}.
\end{align*}
Applying Lemma \ref{lem.minmax.SM}, we obtain the  desired result. 
\end{proof}

\subsection{Proof of Corollary \ref{col.capped}}
\label{sec.col.1}
\begin{proof}
The proof is mostly the same as the proof of Theorem~\ref{thm.stein.dis}. However we need to prove a different version of Lemma \ref{lem.minmax.SM} for the capped $\mathrm{Lip}^L_0(V)$ family. 

Lemma~\ref{lem.minmax.SM} states $g_k(\boldx) \le L\cdot \min_{\boldz\in \partial V}\mathrm{dist}(\boldx, \boldz)=g_0(\boldx)$ for all $g_k \in \mathrm{Lip}^L_0(V)$. Further, we know that $g_k \le 1$ by definition. 
Therefore $g_k(\boldx) \le \min(1,Lg_0(\boldx)) = \bar{g}_L$.  
Notice that $\bar{g}_L \in \overline{\mathrm{Lip}}^L_0(V)$. Using the same argument in Section \ref{sec.minmax.SM}, we can see
\begin{align*}
    \max_{g_k \in \overline{\mathrm{Lip}}^L_0(V)} \mathbbE_q \left[ \sum_k (\partial_k \ell_\boldtheta - \partial_k \logq)^2 g_k\right] 
    = \mathrm{FH}_{\bar{g}_L}(q,p_\boldtheta).
\end{align*}
Applying this result to the last step in Section \ref{appendix:TruncSM_MSD}, gives the desired result. 
\end{proof}

\section{Proof of Theorem \ref{theorem:convergence_GSM}}
\label{eqn:weightfunc}

\subsection{Proof of Theorem~\ref{theorem:convergence_GSM}}
\label{appendix:convergence_GSM}
We assume that 
\begin{align}
 \label{eqn:exp-sup}
 \mathbb{E}\big[ \sup_{\boldtheta: \|\boldtheta-\boldtheta^*\|\leq{\delta}}  |\mathbb{G}_n(m_\boldtheta-m_{\boldtheta^*})|\big]
 \leq C_{\boldg}'\delta^\beta, 
\end{align}
 where $\mathbb{G}_n f=\frac{1}{\sqrt{n}}\sum_{i=1}^{n}(f(X_i)-\mathbb{E}_q[f(X)])$
 for the independent random variables $X_1,\ldots,X_n$ from $q$, and 
 $C_{\boldg}'$ is a positive constant depending on the function $\boldg$
 satisfying the same property as $C_{\boldg}$. 
Then, the estimation accuracy is given by the following theorem. 
\begin{mytheorem}
 [Theorem 5.52 in \citet{vaart00:_asymp_statis}]
 \label{prop:par-accuracy}
 Suppose Assumption \ref{ass.sep},  \ref{ass.thetaconsistency}
 and \eqref{eqn:exp-sup} hold for $\alpha>\beta>0$ and $\alpha>1$. 
 Then, for any positive integer $K$ we have 
 \begin{align*}
  P\left\{\|\hat{\boldtheta}-\boldtheta^*\|> \frac{2^K}{n^{1/(2(\alpha-\beta))}} \right\}
  \leq 
    2^{K(\beta-\alpha)}\frac{2^{2\alpha}}{2^{\alpha-1}-1} \frac{C_{\boldg}'}{C_{\boldg}}. 
 \end{align*}
\end{mytheorem}
Hence, we have $\|\hat{\boldtheta}-\boldtheta^*\|=O_p(n^{-1/(2(\alpha-\beta))})$. 
Our concern is the relation between the coefficient of the convergence rate and the weight function $\boldg$.

The upper bound of the expectation \eqref{eqn:exp-sup} is closely 
related to the covering number of the parameter space $\Theta\subset\mathbb{R}^r$. 
Let us define  $N_{[]}(\varepsilon,\mathcal{F},L^2(Q))$ be the bracketing number of 
$\mathcal{F}$ with the radius $\varepsilon$ under the norm of $L^2(Q)$ and $J_{[]}(\delta,\mathcal{F},L^2(Q))$ be
\begin{align*}
 J_{[]}(\delta,\mathcal{F},L^2(Q))= \int_0^\delta \sqrt{\log N_{[]}(\varepsilon,\mathcal{F},L^2(Q))}\mathrm{d}\varepsilon
\end{align*}
Then, the following theorem holds. 
\begin{mytheorem}
[Corollary 19.35 of \citet{vaart00:_asymp_statis}]
 Let $F$ be an envelope for the function class $\mathcal{F}\subset L^2(Q)$, i.e., 
 $\sup_{f\in\mathcal{F}}\|f(\boldx)\|_{\infty}\leq F(\boldx)$ for every $\boldx\in{V}$ and suppose 
 $\mathbb{E}_{{Q}}[|F(X)|^2]<\infty$. 
 Then, we have 
 \begin{align*}
  \mathbb{E}\big[\sup_{f\in\mathcal{F}}|\mathbb{G}_n(f)|\big]  \leq C \, J_{[]}(\|F\|_{L^2(Q)},\mathcal{F},L^2(Q)), 
 \end{align*}
where $C$ is a universal constant. 
\end{mytheorem}
The bracketing number of the parametric model is given by the following proposition. 
\begin{prop} 
[Example 19.6 in \citet{vaart00:_asymp_statis}]
 \label{prop:para-bracket}
 The bracketing number of the parametrized loss function $m_\boldtheta(\boldx)$ is given as follows. 
 Let $\Theta\subset\mathrm{R}^r$  be contained in a ball of radius $R$. 
 Let $\mathcal{F}=\{m_\boldtheta(\boldx)\,|\,\boldtheta\in\Theta\}$ be a function class indexed by $\Theta$. 
 Suppose there exists a function $\dot{m}(\boldx)$ with $\|\dot{m}\|_{L^2(Q)}<\infty$ such that
 \begin{align*}
  |m_{\boldtheta_1}(\boldx)-m_{\boldtheta_2}(\boldx)| \leq \dot{m}(\boldx)\|\boldtheta_1-\boldtheta_2\|_2
 \end{align*}
 for all $\boldx\in{V}$ and $\boldtheta_1,\boldtheta_2\in\Theta$. Then, for every $\varepsilon>0$, 
 \begin{align*}
  N_{[]}(\varepsilon\|\dot{m}\|_{L^2(Q)},\mathcal{F},L^2(Q))
  \leq 
  \left(1+\frac{4R}{\varepsilon}\right)^r.
 \end{align*} 
\end{prop}

In our case, we need to evaluate $\mathbb{E}[\sup_{f\in\mathcal{F}_\delta} |\mathbb{G}_n(f)|]$ for 
\begin{align*}
 \mathcal{F}_\delta:=\{m_{\boldtheta}(\boldx)-m_{\boldtheta^*}(\boldx)
 \,|\,\boldtheta\in\Theta,\,\|\boldtheta-\boldtheta^*\|\leq \delta\}, 
\end{align*}
where $\boldtheta^*$ is the minimizer of the expected loss $M(\boldtheta)$. 
The function $\dot{m}(\boldx)$ in Proposition~\ref{prop:para-bracket} should satisfy 
\begin{align*}
 |(m_{\boldtheta_1}(\boldx)-m_{\boldtheta^*}(\boldx)) - 
 (m_{\boldtheta_2}(\boldx)-m_{\boldtheta^*}(\boldx))| 
 =
 |m_{\boldtheta_1}(\boldx) - m_{\boldtheta_2}(\boldx)|  
 \leq  \dot{m}(\boldx)\|\boldtheta_1-\boldtheta_2\|. 
\end{align*}
Then, Proposition~\ref{prop:para-bracket} leads to 
\begin{align*}
  N_{[]}(\varepsilon\|\dot{m}\|_{L^2(Q)}, \mathcal{F}_\delta, L^2(Q))
 \leq 
  \left(1+\frac{4\delta}{\varepsilon}\right)^r. 
\end{align*}
The envelope function $F_\delta(\boldx)$ of $\mathcal{F}_\delta$ is given by $F_\delta(\boldx)=\dot{m}(\boldx)\delta$, because 
\begin{align*}
 |m_{\boldtheta}(\boldx)-m_{\boldtheta^*}(\boldx)|\leq 
 \dot{m}(\boldx)\|\boldtheta-\boldtheta^*\|\leq \dot{m}(\boldx)\delta. 
\end{align*}
Hence, we obtain
\begin{align*}
 N_{[]}(\varepsilon\|F_\delta\|_{L^2(Q)},\mathcal{F}_\delta,L^2(Q))
 =
 N_{[]}(\varepsilon \delta \|\dot{m}\|_{L^2(Q)},\mathcal{F}_\delta,L^2(Q))
 \leq 
\left(1+\frac{4}{\varepsilon}\right)^r, 
\end{align*}
and thus, 
\begin{align*}
 \mathbb{E}[\sup_{f\in\mathcal{F}_\delta} |\mathbb{G}_n(f)|]
 \leq
 C\delta\|\dot{m}\|_{L^2(Q)}\sqrt{r}
 \int_0^1 \sqrt{\log\bigg(1+\frac{4}{\varepsilon}\bigg)}\mathrm{d}\varepsilon  
 \leq
 C'\sqrt{r}\delta\|\dot{m}\|_{L^2(Q)}
\end{align*}
holds, where $C$ and $C'$ are universal constants. We find that $\beta$ in \eqref{eqn:exp-sup} 
is given by $\beta=1$. 

Let us evaluate the norm $\|\dot{m}\|_{L^2(Q)}$. 
For the function $A_k$ and $B_k$ in \eqref{eqn:estfunc_TruncSM}, we have the following inequalities
\begin{align}
\begin{array}{l}
 |A_k(\boldx,\boldtheta_1)-A_k(\boldx,\boldtheta_2)| \leq 
  \dot{A}_k(\boldx)\|\boldtheta_1-\boldtheta_2\|,\\
 |B_k(\boldx,\boldtheta_1)-B_k(\boldx,\boldtheta_2)| \leq 
 \dot{B}_k(\boldx)\|\boldtheta_1-\boldtheta_2\|. 
\end{array}
\end{align}
It is straightforward to see that the following $\dot{m}$ satisfies the required condition: 
\begin{align*}
 \dot{m}(\boldx) = 
 \sum_k\{ g_k(\boldx) \dot{A}_k(\boldx) +  |\partial_k g_k(\boldx)| \dot{B}_k(\boldx)\}. 
 \end{align*}
Cauchy-Schwarz inequality leads to 
$\|g_k\dot{A}_k \|_{L^2(Q)} 
 \leq   (\mathbb{E}_{q}[\dot{A}_k^4])^{1/4} (\mathbb{E}_{q}[g_k^4])^{1/4}$. 
The similar inequality holds for $\|\dot{B}_k \partial_{k}g_k\|_{L^2(Q)}$. 
Hence, we have 
\begin{align}
\label{eqn:Gamma_g}
 \|\dot{m}\|_{L^2(Q)} 
 &\leq 
 \Gamma(\boldg;A,B):=
 \sum_{k=1}^{d}  
    \bigg\{
    (\mathbb{E}_{q}[\dot{A}_k^4]\mathbb{E}_{q}[g_k^4])^{1/4}
 +  (\mathbb{E}_{q}[\dot{B}_k^4]\mathbb{E}_{q}[|\partial_{k}g_k|^4])^{1/4}3
  \bigg\}. 
\end{align}
In summary, we have the estimation error bound in Theorem~\ref{theorem:convergence_GSM}. 

\section{Some More Analysis of Weight Functions}
\label{sec.capped.distance}

In this subsection, we assume that 
the statistical model is realizable, i.e., $q=p_{\boldtheta^*}$ holds. 
Let $U$ be a subset in the domain $V$. 
Under the regularity condition later shown in Appendix~\ref{appendix:deriv_Cg}, 
one can find that the constant $C_{\boldg}$ in~\eqref{eqn:ass1} is given by 
\begin{align}
\label{eqn:C_ming}
C_{\boldg}=\min_{\boldx\in{U}}\min_{k\in[d]} g_k(\boldx) 
\end{align}
up to a constant independent of $\boldg$. 
As the continuous and non-negative weight function $g_k(\boldx)$ can take zero only on the boundary $\partial{V}$, 
$C_{\boldg}>0$ holds if $U$ is a closed subset which does not include boundary points of $V$.

Let us consider $\Gamma(\boldg;A,B)/C_{\boldg}$ in the upper bound in 
Theorem~\ref{theorem:convergence_GSM}. 
The following theorem ensures that the minimum function 
\begin{align}
\label{eqn:minfunc}
 h(\boldx) = \min_{k\in[d]}g_k(\boldx)
\end{align}
 improves the upper bound of the estimation error under some conditions on $g_k$. 
 \begin{mytheorem}
\label{thm:min_weight}
 Let $\boldg=(g_1,\ldots,g_d)$  and $\boldh=(h,\ldots,h)$ 
 be the weight functions, where $h$ is defined by \eqref{eqn:minfunc}. 
 Suppose that $g_k$ is differentiable on $V$ except a measure zero set
 and that the set 
 $\{\boldx\in{V}\,|\,g_k(\boldx)=g_{k'}(\boldx),\,\exists k,k'\in[d], k\neq{k}'\}$ 
  is measure zero. We assume that 
\begin{align*}
|\partial_k g_k(\boldx)|\geq |\partial_k g_{k^*}(\boldx)|, 
\end{align*} 
holds for $\boldx\in{V}$ and $k\in[d]$, where $k^*\in[d]$ is the number such that $h(\boldx)=g_{k^*}(\boldx)$. 
Then, we have
\begin{align*}
    \frac{\Gamma(\boldh;A,B)}{C_{\boldh}}
    \leq
    \frac{\Gamma(\boldg;A,B)}{C_{\boldg}}, 
\end{align*}
where $C_{\boldg}$ and $C_{\boldh}$ are defined by 
\eqref{eqn:C_ming}. 
\end{mytheorem}
The proof is found in Appendix~\ref{appendix:Proof_thm_minweight}. 

\begin{exam}[Bounded Rectangular Domain]
For the $d$-dimensional rectangular $V=\prod_{k=1}^d[0,c_k]$, 
let us consider the statistical accuracy of the generalized score matching (SM) method~\citep{YuGen2019}
with $g_k(\boldx)=\min\{x_k, c_k-x_k\}$, and $h(\boldx)=\min_{k}g_k(\boldx)$. 
Note that the weight $h$ is nothing but $g_0$ in \eqref{eqn:g0_distance_func}. 
According to Theorem~\ref{thm:min_weight}, we find that
the estimator with the weight $h$ is superior to 
the estimator with the above $g_1,\ldots,g_d$ 
in the sense of the estimation error bound. 
\end{exam}

\begin{exam}[Unit ball under $p$-norm]
\label{exam:unit-p-ball}
Let us define $V$ as the $d$-dimensional unit ball under $p$-norm, $V=\{\boldx\in\mathbb{R}^d|\|\boldx\|_p\leq 1\}$. 
The weight $g_k(\boldx)$ is defined by the distance from $\boldx$ to the boundary of $V$ along the $k$-th axis. 
This is expressed by 
\begin{align*}
g_k(\boldx) = \max\{|\varepsilon|\,|\,\boldx+\varepsilon\bolde_k\in V\} = 
 \min\{(1-\|\boldx_{(k)}\|_p^p)^{1/p}-x_k,\, x_k+(1-\|\boldx_{(k)}\|_p^p)^{1/p}\}, 
\end{align*}
where $\bolde_k$ is the unit vector along with the $k$-th axis, 
$\boldx_{(k)}$ is the $d-1$ dimensional vector dropped the $k$-th element $x_k$ from $\boldx$. 
Some calculation yields the inequality 
$|\partial_k g_k(\boldx)|=1\geq |\partial_k g_{\ell}(\boldx)|$ for $\ell=\argmin_{k}g_k(\boldx)$. 
Hence, the minimum function $h(\boldx)=\min_{k}g_k(\boldx)$ improves the error bound of the estimator with the weight $g_k$. 
One can confirm that $h(\boldx)=\min_{\boldz\in\partial{V}}\|\boldx-\boldz\|_1$ holds. 
Likewise, 
for the truncated domain 
$V=\{\boldx=(x_1,\ldots,x_d)\in\mathbb{R}^d|\|\boldx\|_p\leq 1,\, x_i\ge c_i\},\, c_i\in \mathbb{R}$, one can prove the inequality 
$|\partial_k g_k(\boldx)|=1\geq |\partial_k g_{\ell}(\boldx)|$. 
The above assertions are explained in Appendix~\ref{Appendix:unit-p-ball}. 
\end{exam}

Under the assumption of Theorem~\ref{thm:min_weight}, we find that the homogeneous weight $\boldh$ improves the upper bound of the estimation error for the estimator with $\boldg$. 
Furthermore, the weight $\boldh$ corresponds to the minimum $\ell^1$ distance to the boundary 
as shown in Example~\ref{exam:unit-p-ball}. 
To compare the distance-based homogeneous weights, we need to evaluate the upper bound for each distance. 
In Appendix~\ref{appendix:ell1_ell2}, 
we show numerical comparison between $\ell^1$ and $\ell^2$ distances and numerical evaluation of the theoretical upper bound.

\subsection{Derivation of (\ref{eqn:C_ming})}
\label{appendix:deriv_Cg}
Suppose the true probability $q(\boldx)$ is $p_{\boldtheta^*}(\boldx)$. 
Let $U$ be an open subset of $V$. Then, we have 
\begin{align*}
 M(\boldtheta)-M(\boldtheta^*) 
 &=
 \int_V\sum_k g_k(\boldx) (\partial_k \log{p_\boldtheta(\boldx)}-\partial_k
 \log{p_{\boldtheta^*}(\boldx)})^2 p_{\boldtheta^*}(\boldx) d\boldx \\
 &\geq
  \int_U\sum_k g_k(\boldx) (\partial_k \log{p_\boldtheta(\boldx)}-\partial_k
 \log{p_{\boldtheta^*}(\boldx)})^2 p_{\boldtheta^*}(\boldx) d\boldx \\
 &\geq 
 \min_{\boldx\in{U}}\min_k\{g_k(\boldx)\}
 \int_{U}\sum_k (\partial_k \log{p_\boldtheta(\boldx)}-\partial_k \log{p_{\boldtheta^*}(\boldx)})^2
 p_{\boldtheta^*}(\boldx) d\boldx. 
\end{align*}
Suppose that the Hessian matrix of 
\begin{align}
\int_{U}\sum_k (\partial_k \log{p_\boldtheta(\boldx)}-\partial_k \log{p_{\boldtheta^*}(\boldx)})^2
 p_{\boldtheta^*}(\boldx) d\boldx    
\end{align}
as the function of $\boldtheta$ is non-degenerate, 
there exists a constant $C_0$ independent of $\boldg$ such that 
the above integral is bounded below by $C_0\|\boldtheta-\boldtheta^*\|^2$. 
Hence, $C_{\boldg}= \min_{\boldx\in{U}}\min_k\{g_k(\boldx)\}$ satisfies 
the required condition.

\subsection{Proof of Theorem~\ref{thm:min_weight}}
\label{appendix:Proof_thm_minweight}
From the definition of $h$, one can find 
\begin{align*}
 C_{\boldg} = \min_{\boldx\in{U}}\min_{k} 
 g_k(\boldx) = \min_{\boldx\in{U}}h(\boldx) = C_{\boldh}. 
\end{align*}
We evaluate $\Gamma(\boldg;A,B)$ and $\Gamma(\boldh;A,B)$. 
As for the derivative function, the assumption guarantees that
$\partial_k h(\boldx) =  \partial_k g_{k^*}(\boldx)$ and 
$|\partial_k g_k(\boldx)|\geq |\partial_k g_{k^*}(\boldx)|$. 
Hence, we have
\begin{align*}
 \Gamma(\boldg;A,B)
 &=
 \sum_{k=1}^{d}  
    (\mathbb{E}_{q}[\dot{A}_k^4]\mathbb{E}_{q}[g_k^4])^{1/4}
 + 
 \sum_{k=1}^{d}   (\mathbb{E}_{q}[\dot{B}_k^4]\mathbb{E}_{q}[|\partial_{k}g_k|^4])^{1/4} \\
&\ge  
 \sum_{k=1}^{d}  
 (\mathbb{E}_{q}[\dot{A}_k^4]\mathbb{E}_{q}[(\min_{k}g_k)^4])^{1/4}
 + 
 \sum_{k=1}^{d}
 (\mathbb{E}_{q}[\dot{B}_k^4]\mathbb{E}_{q}[|\partial_{k}g_{k^*}|^4])^{1/4}\\
& =\Gamma(\boldh;A,B). 
\end{align*}
As a result, we obtain $\frac{\Gamma(\boldg;A,B)}{C_{\boldg}}\ge \frac{\Gamma(\boldh;A,B)}{C_{\boldh}}$.

\subsection{Some Equations in Example~\ref{exam:unit-p-ball}}
\label{Appendix:unit-p-ball}
\textbf{Proof of $|\partial_k g_k(\boldx)| \ge |\partial_k{g_{k^*}}(\boldx)|$:} 
 Without loss of generality, we suppose $\boldx=(x_1,\ldots,x_d)\geq \bold0$. Then, $g_k(\boldx) = (1-\|\boldx_{(k)}\|_p^p)^{1/p}-x_k$ holds. 
 Firstly, we prove that $g_\ell(\boldx) < g_k(\boldx)$ leads to $x_k < x_\ell$. For the sake of simplicity let us 
 assume $k=1$ and $\ell=2$. 
 Let us define $c^p = 1-x_3^p-\cdots-x_d^p$ for $c\geq 0$. Then, 
 we find that $g_2(\boldx) < g_1(\boldx)$ leads to $(c^p-x_1^p)^{1/p}+x_1 < (c^p-x_2^p)^{1/p}+x_2$. 
 The function $f(x)=(c^p-x^p)^{1/p}+x$ for $0\leq x\leq c$ is concave, takes the maximum value at
 $x=c/2^{1/p}(<c)$ and satisfies $f(x)=f((c^p-x^p)^{1/p})$. 
 When $y$ lies on the interval between $x$ and $(c^p-x^p)^{1/p}$, 
 $f(x)=f((c^p-x^p)^{1/p})\leq f(y)$ holds. 
 Suppose $x_1^p+x_2^p\leq c^2$ and $f(x_1)<f(x_2)=f((c^p-x_2^p)^{1/p})$. 
 If $x_1$ lies on the interval between 
 $x_2$ and $(c^p-x_2^p)^{1/p}$, $f(x_2)\leq f(x_1)$ holds and it is the contradiction. 
 Hence we have $x_1 < x_2$. 
 Next, we evaluate the absolute value of the derivatives. When  $g_\ell(\boldx)< g_k(\boldx)$, $x_k<x_\ell$ holds. 
 For $k\neq{\ell}$, we obtain $|\partial_k g_k(\boldx)|=1$ and 
 $|\partial_k g_\ell(\boldx)|=|x_k|^{p-1}(1-\|\boldx_{(\ell)}\|_p^p)^{1/p-1} \leq|x_\ell|^{p-1}(|x_\ell|^{p})^{1/p-1}=1$ for $p\geq 1$. 

\textbf{Rough sketch of the proof of $h(\boldx)=\min_{\boldz\in\partial{V}}\|\boldx-\boldz\|_1$:} 
Suppose that all the vertexes of the polytope $B_1(\boldx,c):=\{\boldz\in\mathbb{R}^d|\|\boldx-\boldz\|_1\leq c\}$ are included in $V$. 
Then, $B_1(\boldx,c)\subset{V}$ holds as $V$ is convex. Clearly, $h(\boldx)$ is expressed by 
$\sup_{c\geq 0}\{c|B_1(\boldx,c)\subset{V}\}$. This is nothing but the $\ell^1$-distance from $\boldx$ to the boundary of $V$.

\section{Derivation of (\ref{eqn:lower-upper-Boundary})}
\label{Appendix:derivation_lower-upper-BoundaryCond}
Let us define $U$ by  
$U=\{\boldx\in{V}|g_0(\boldx)\ge{c}\}$.
We see that $\min_{\boldx\in{U}}\bar{g}_{L}(\boldx)=1$ holds from Section \ref{appendix:deriv_Cg}. 
Hence we have $C_{\bar{g}_L}=1$. 
Let us evaluate the fourth order moments of $\bar{g}_L$: %
\begin{align*}
\mathbb{E}[\bar{g}_{L}^4]
&=
L^4\int_{0}^{1/L} z^4 p_{\mathrm{dist}}(z)dz
+
\int_{z\ge 1/L} p_{\mathrm{dist}}(z)dz \\
&=
L^4\int_{0}^{1/L} z^4 p_{\mathrm{dist}}(z)dz
+
1-\int_{0}^{1/L} p_{\mathrm{dist}}(z)dz
=1-\frac{C}{L^{1+\beta}}, 
\end{align*}
where $C$ is a positive constant such as 
$c_{b,\beta}:=\frac{4 b} {\beta^2+6 \beta+5}\le C\le C_{b',\beta}:=\frac{4 b'}{\beta^2+6 \beta+5}$. 
Due to the non-negativity of $\mathbb{E}[\bar{g}_{L}^4]$, 
$L^{1+\beta}\ge C$ should hold. This inequality is guaranteed from 
$1\ge \int_{0}^{1/L}b' z^{\beta}dz\ge \int_{0}^{1/L}p_{\mathrm{dist}}(z)dz$. 
Let us evaluate the second term of $\Gamma(\boldg;A,B)$. 
The derivative $\partial_k\bar{g}_{L}$ is given by 
\begin{align*}
\partial_k\bar{g}_{L}(\boldx)=
\begin{cases}
\displaystyle
L\frac{x_k-\widetilde{x}_k}{\|\boldx-\widetilde{\boldx}\|},&  g_0(\boldx)<1/L,\\
\displaystyle
\ 0,& g_0(\boldx)>1/L, 
\end{cases}
\end{align*}
where $\widetilde{\boldx}$ is the minimum solution of 
$\min_{\boldz\in\partial{V}}\|\boldx-\boldz\|$. Hence, we have
\begin{align*}
\mathbb{E}[|\partial_k\bar{g}_{L}|^4] 
&= 
L^4\int_{g_0(\boldx)\le 1/L}\frac{|x_k-\widetilde{x}_k|^4}{g_0(\boldx)^4} q(\boldx)d\boldx 
\le
L^4\int_{0}^{1/L}p_{\mathrm{dist}}(z)dz
\le \frac{b'}{\beta+1} L^{-\beta+3}. 
\end{align*}
Next, we evaluate the lower bound of the fourth moment 
of the derivative $\partial_k g_k$. Jensen's inequality yields that 
\begin{align*}
\mathbb{E}[|\partial_k\bar{g}_{L}|^4]
&=
L^4\sum_{k=1}^{d}  
\int_{g_0(\boldx)\le 1/L}\frac{|x_k-\widetilde{x}_k|^4}{g_0(\boldx)^4} q(\boldx)d\boldx \\
&\ge
d L^4 
\left(
\frac{1}{d}\sum_{k=1}^{d}
\int_{g_0(\boldx)\le 1/L}\frac{|x_k-\widetilde{x}_k|^2}{g_0(\boldx)^2} q(\boldx)d\boldx \right)^2 \\
&=
\frac{L^4}{d}
\left(\int_{g_0(\boldx)\le 1/L}q(\boldx)d\boldx \right)^2 
\ge 
\frac{b^2}{d(\beta+1)^2} L^{2(1-\beta)}
\end{align*}
Let us define $C_{A}=\sum_{k=1}^{d}(\mathbb{E}_q[\dot{A}_k^4])^{1/4},
C_{B}=\sum_{k=1}^{d}(\mathbb{E}_q[\dot{B}_k^4])^{1/4},
c_0=(b^2/(d(\beta+1)^2))^{1/4}$, and $c_1=(b'/(\beta+1))^{1/4}$. 
Then, we have 
\begin{align*}
C_{A}\left(1-\frac{C_{b',\beta}}{L^{\beta+1}}\right)^{1/4}+C_{B}c_0 L^{(1-\beta)/2}
\le
\frac{\Gamma(\bar{g}_L;A,B) }{C_{\bar{g}_L}}
\le
C_{A}\left(1-\frac{c_{b,\beta}}{L^{\beta+1}}\right)^{1/4}+C_{B}c_1  L^{(3-\beta)/4}. 
\end{align*}

\section{Choice of Weight Function: $\ell^1$  vs. $\ell^2$ distance}
\label{appendix:ell1_ell2}

\begin{figure}
    \centering
    \includegraphics[width = .45\textwidth]{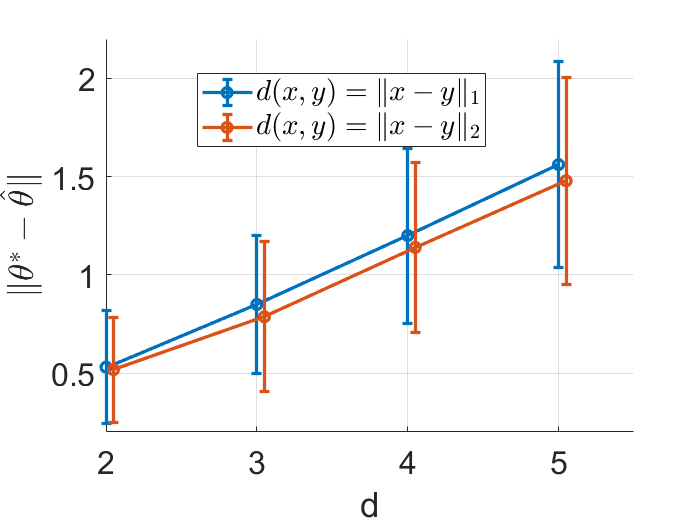}
    \caption{Estimation accuracy of TruncSM using $\ell^1$ vs. $\ell^2$ distance.}
    \label{fig:l1vl2}
\end{figure}

\begin{table}[h]\centering

\caption{Comparison of the weight $h_p,\, p=1,2$. The table shows 
$\big\{R d \mathbb{E}_{q}[|h_p|^4]^{1/4} + (1-R)\sum_{k=1}^{d}\mathbb{E}_{q}[|\partial_k h_p|^4]^{1/4}\big\}/C_{\boldh_p}$ for $R=0.2, 0.8$. }
\label{tbl:L1L2comp}\small 
\begin{tabular}{c|ccccc|ccccc}
\multicolumn{1}{c}{} &  \multicolumn{5}{|c}{$R=0.2$} &    \multicolumn{5}{|c}{$R=0.8$}   \\\hline
 weight $\backslash$ $\dim$              &   2    &  3    &   4   &  5   &   6    &   2  &   3  &   4   &  5   &   6   \\ \hline
$h_1$                  &  3.160 &4.558  &5.939  &7.309 & 8.685  & 1.569 &2.088& 2.537 &2.975 &3.371 \\
$h_2$                  &  3.073 &4.279  &5.412  &6.496 & 7.552  & 1.545 &2.012& 2.402 &2.764 &3.081 \\
\end{tabular}
\end{table} 

In this section, we test different metrics for $g_0$: $\ell^1$~distance $\mathrm{dist}(\boldx,\boldy):= \|\boldx-\boldy\|_1$ and $\ell^2$~distance  $\mathrm{dist}(\boldx,\boldy):= \|\boldx-\boldy\|_2$. The data set is generated from $\mathcal{N}(\boldone_d\cdot 0.5, \boldI_d )$, where $\boldone_d$ and $\boldI_d$ are the $d$ dimensional all one vector and the $d\times d$ identity matrix respectively. The truncation domain is $V = \{\boldx | \|\boldx\|_1 < 1 \text{ and } x_d > 0\}$. 150 samples are generated. We plot the estimation error $\|\hat{\boldtheta} - \boldtheta^*\|$ and its standard deviation for both choices of distances: $\ell^1$ and $\ell^2$ versus the dimensionality $d$ in Figure \ref{fig:l1vl2}. It can be seen that the Euclidean distance, i.e., $\ell^2$ distance consistently achieves slightly lower error than the $\ell^1$ distance.  

Let us numerically evaluate the upper bound $\Gamma(\boldh_p;A,B)/C_{\boldh_p}$ 
for the weight $\boldh_p=(h_p,\ldots,h_p)$, where 
$h_p(\boldx)=\min_{\boldz\in\partial{V}}\|\boldx-\boldz\|_p,\, p=1,2$. 
Let ``$\sup_{U}$'' be the supremum over all open subsets of $V$. Then, we have 
$\sup_{U}\min_{\boldx\in{U}}h_1(\boldx)=1/2$ and $\sup_{U}\min_{\boldx\in{U}}h_2(\boldx)=1/(\sqrt{d}+1)$. 
In order to obtain the tight bound, we use $C_{\boldh_1}=1/2$ and $C_{\boldh_2}=1/(\sqrt{d}+1)$ for our evaluation. For $\bar{A}=\max_k\mathbb{E}_q[|\dot{A}_k|^4]^{1/4}, \bar{B}=\max_k\mathbb{E}_q[|\dot{B}_k|^4]^{1/4}$,
and $R=\bar{A}/(\bar{A}+\bar{B})$, clearly we have 
\begin{align*}
\Gamma(\boldh_p;A,B) 
 \leq 
 (\bar{A}+\bar{B})\big\{R d \mathbb{E}_{q}[|h_p|^4]^{1/4} + (1-R)\sum_{k=1}^{d}\mathbb{E}_{q}[|\partial_k h_p|^4]^{1/4}\big\}. 
\end{align*}
For $\boldx\sim N_d(\bold1_d\cdot 0.5, \boldI_d)$, we numerically evaluate $\mathbb{E}_{q}[|h_p|^4]$ 
and $\mathbb{E}_{q}[|\partial_k h_p|^4]$. Table~\ref{tbl:L1L2comp} shows 
$\big\{R d \mathbb{E}_{q}[|h_p|^4]^{1/4} + (1-R)\sum_{k=1}^{d}\mathbb{E}_{q}[|\partial_k h_p|^4]^{1/4}\big\}/C_{\boldh_p}$
of each dimension and weight for $R=0.2, 0.8$. 

In this problem setup, one can confirm that the upper bound for $p=2$ is slightly smaller than that of $p=1$ for all
$R\in[0,1]$, as the bound is the linear function of $R$. 
The theoretical conclusion verifies the numerical results. 

\section{Increasing Boundary Size in Section \ref{sec.weightfunc}}
\label{appendix:boundary_size}
When experimenting with different values of $L$ used in capping the distance function $\bar{g}_L$, an increasing boundary size (shown by values of $b$ in Figure \ref{fig:boundary}) is used in the experiments. The two regions, the square boundary and the disjoint boundary, are created by supplying set(s) of vertices $\Omega$, detailing the locations of the polygonal truncation domain. The variable $b$ controls the boundary size by offsetting the supplied vertices in $\Omega$. For the square, this was
\[
\Omega_{\mathrm{sq}} := \{( -b,  -b), ( -b, b), (b, b), (b, -b)\}.
\]
Increasing $b$ in this scenario leads to the corners of the square boundary being shifted by an equal amount.
For the disjoint boundary, two sets of vertices were given for two disjoint domains:
\begin{align*}
\Omega_{\mathrm{dis1}} := &\{ (1-b, 0.5-b), (1-b, 0.5), (1+b, 0.5), (1+b, 0.5-b)\}, \\
\Omega_{\mathrm{dis2}} := &\{(1-b, 1.5), (1-b, 1.5+b),
              (1+b, 1.5+b), (1+b, 1.5)\}.
\end{align*}
Increasing $b$ in this case will enlarge the truncation domain while the ``disjoint'' section in the middle remains unchanged.

\section{Chicago Crime data set with Different Random Initialization}
\label{appendix:Chicago}
\begin{figure}[t]
    \centering
    \includegraphics[width = .5\textwidth]{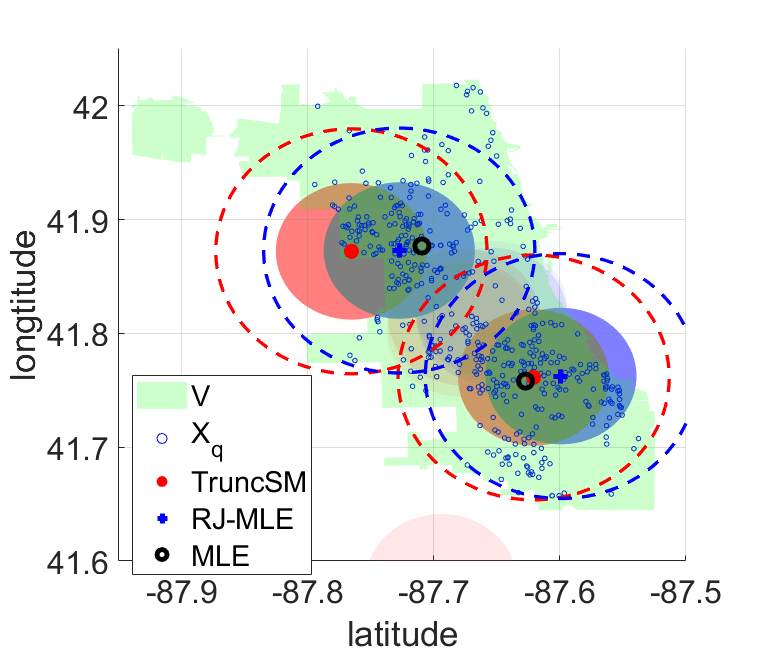}
    \caption{Chicago data set: Gaussian mixture fitting with random initialization.}
    \label{fig:chishade}
\end{figure}

As we mentioned in Section \ref{sec.chi}, both TruncSM and RJ-MLE solve non-convex optimization problems so it is possible for them to return local optima. Therefore, to show the stability of both methods' solutions, in this experiment, we randomly initialize both methods with the initial point $\boldsymbol{\mu}_q + \epsilon, $ where $\boldsymbol{\mu}_q$ is the mean of the data set $X_q$ and $\epsilon  \sim \mathcal{N}(\boldzero, \boldI_d \cdot 0.06^2)$. Both methods were executed 500 times and each time, we plot two lightly shaded balls centered at the estimated mixture centers with their radius equal to the standard deviation. One can see from the blue/red shades in Figure \ref{fig:chishade}, both algorithms consistently place centers at northern and southern Chicago, and both balls are roughly centered at locations reported in the Section \ref{sec.chi}. TruncSM also placed a Gaussian center outside of the boundary of Chicago in one of the simulations. This is a very rare event and such a result can be easily ruled out. 

\end{appendices}

\newpage

\bibliography{references}

\end{document}